\documentclass[12pt]{iopart}

\usepackage{natbib}
\usepackage[utf8]{inputenc} 
\usepackage[T1]{fontenc}    
\usepackage{url}            
\usepackage{booktabs}       
\usepackage{amsfonts}       
\usepackage{nicefrac}       
\usepackage{microtype}      
\usepackage{amssymb}
\usepackage{bm}
\usepackage{subfigure}
\usepackage{eucal}
\usepackage{graphicx}
\usepackage{epsfig}
\usepackage{amssymb}
\expandafter\let\csname equation*\endcsname\relax
\expandafter\let\csname endequation*\endcsname\relax
\usepackage{amsmath}
\usepackage{amsthm}
\usepackage{enumitem}

\usepackage{fancyhdr}

\newtheorem*{remark}{Remark} 
\newtheorem{proposition}{Proposition}

\usepackage{hyperref}
\hypersetup{
    unicode=false,          
    pdftoolbar=true,        
    pdfmenubar=true,        
    pdffitwindow=false,     
    pdfstartview={FitH},    
    pdftitle={My title},    
    pdfauthor={Author},     
    pdfsubject={Subject},   
    pdfcreator={Creator},   
    pdfproducer={Producer}, 
    pdfkeywords={keyword1, key2, key3}, 
    pdfnewwindow=true,      
    colorlinks=true,       
    linkcolor=cyan,          
    citecolor=cyan,        
    filecolor=cyan,         
    urlcolor=cyan        
}

\theoremstyle{definition}
\newtheorem{defn}{Definition} 
\DeclareMathAlphabet{\mathcal}{OMS}{cmsy}{m}{n} 

\setcitestyle{numbers,open={[},close={]},citesep={,}}

\begin{document}
\title[Robust Bayesian Optimisation]{Achieving Robustness to Aleatoric Uncertainty with Heteroscedastic Bayesian Optimisation}

\author{Ryan-Rhys Griffiths$^1$, Alexander A Aldrick$^1$, \\Miguel Garcia-Ortegon$^{2, 3}$, Vidhi Lalchand$^2$ and Alpha A. Lee$^1$}\address{$^1$ Department of Physics, University of Cambridge}\address{$^2$ Department of Engineering, University of Cambridge}\address{$^3$ Department of Mathematics University of Cambridge} \ead{rrg27@cam.ac.uk}

\begin{abstract}
    Bayesian optimisation is a sample-efficient search methodology that holds great promise for accelerating drug and materials discovery programs. A frequently-overlooked modelling consideration in Bayesian optimisation strategies however, is the representation of heteroscedastic aleatoric uncertainty. In many practical applications it is desirable to identify inputs with low aleatoric noise, an example of which might be a material composition which displays robust properties in response to a noisy fabrication process. In this paper, we propose a heteroscedastic Bayesian optimisation scheme capable of representing and minimising aleatoric noise across the input space. Our scheme employs a heteroscedastic Gaussian process (GP) surrogate model in conjunction with two straightforward adaptations of existing acquisition functions. First, we extend the augmented expected improvement (AEI) heuristic to the heteroscedastic setting and second, we introduce the aleatoric noise-penalised expected improvement (ANPEI) heuristic. Both methodologies are capable of penalising aleatoric noise in the suggestions. In particular, the ANPEI acquisition yields improved performance relative to homoscedastic Bayesian optimisation and random sampling on toy problems as well as on two real-world scientific datasets. Code is available at: \url{https://github.com/Ryan-Rhys/Heteroscedastic-BO}
    
\end{abstract}


\maketitle

\section{Introduction}

Bayesian optimisation is proving to be a highly effective search methodology in areas such as drug discovery \citep{2018_Design, 2020_Griffiths, 2020_Hoffman}, materials discovery \cite{2020_Hase, 2020_Olympus, 2020_Terayama}, chemical reaction optimisation \cite{2020_Felton, 2020_Summit, 2020_Yehia}, robotics \cite{2016_Calandra}, sensor placement \cite{2019_Grant}, tissue engineering \cite{2018_Olofsson} and genetics \cite{2020_Moss_Boss}. Heteroscedastic aleatoric noise however, is rarely accounted for in these settings despite being an important consideration for real-world applications. Aleatoric uncertainty refers to uncertainty inherent in the observations (measurement noise) \cite{2017_Kendall}. In contrast, epistemic uncertainty corresponds to model uncertainty and may be explained away given sufficient data. Heteroscedastic aleatoric noise refers to aleatoric noise which varies across the input domain and is a prevalent feature of many scientific datasets; perhaps suprisingly not only experimental datasets, but also datasets where properties are predicted computationally. One such source of heteroscedasticity in the computational case might be situations in which the accuracy of first-principles calculations deteriorate as a function of the chemical complexity of the molecule being studied \cite{2018_Griffiths}. 

\begin{figure*}[!ht]
\centering
\subfigure[Density plot of computational errors]{\label{fig:hist_DFT}\includegraphics[width=0.48\textwidth]{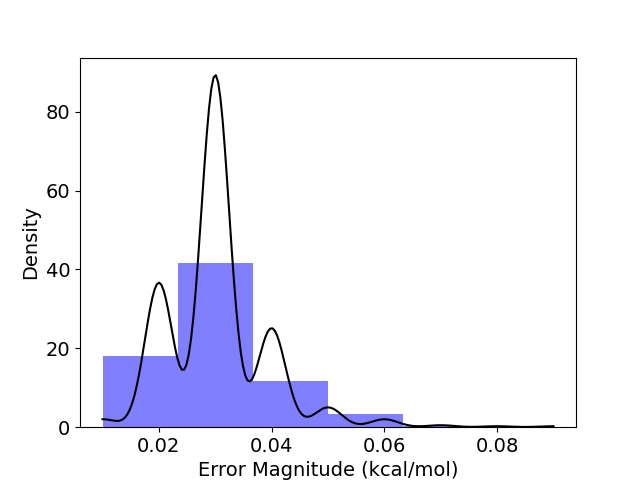}}
\subfigure[Density plot of experimental errors]{\label{fig:hist_exp}\includegraphics[width=0.48\textwidth]{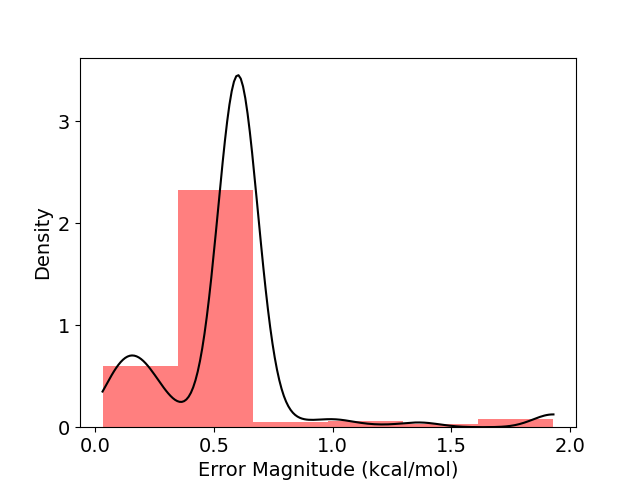}}
\caption{(a) The density histogram of computational errors (kcal/mol) for the FreeSolv hydration energy dataset (\cite{2017_Duarte}). The computational errors in the hydration free energy arise from systematic errors in the force field used in alchemical free energy calculations based on classical molecular dynamics (MD) simulations. (b) A similar density histogram for the experimental errors where the source of uncertainty stems from the instrumentation used to obtain the measurement. The histograms are overlaid with kernel density estimates}.
\label{fig:hist}
\end{figure*}

\begin{figure*}[ht!]
\centering
\subfigure[Homoscedastic GP Fit ]{\label{fig:homo}\includegraphics[width=0.48\textwidth]{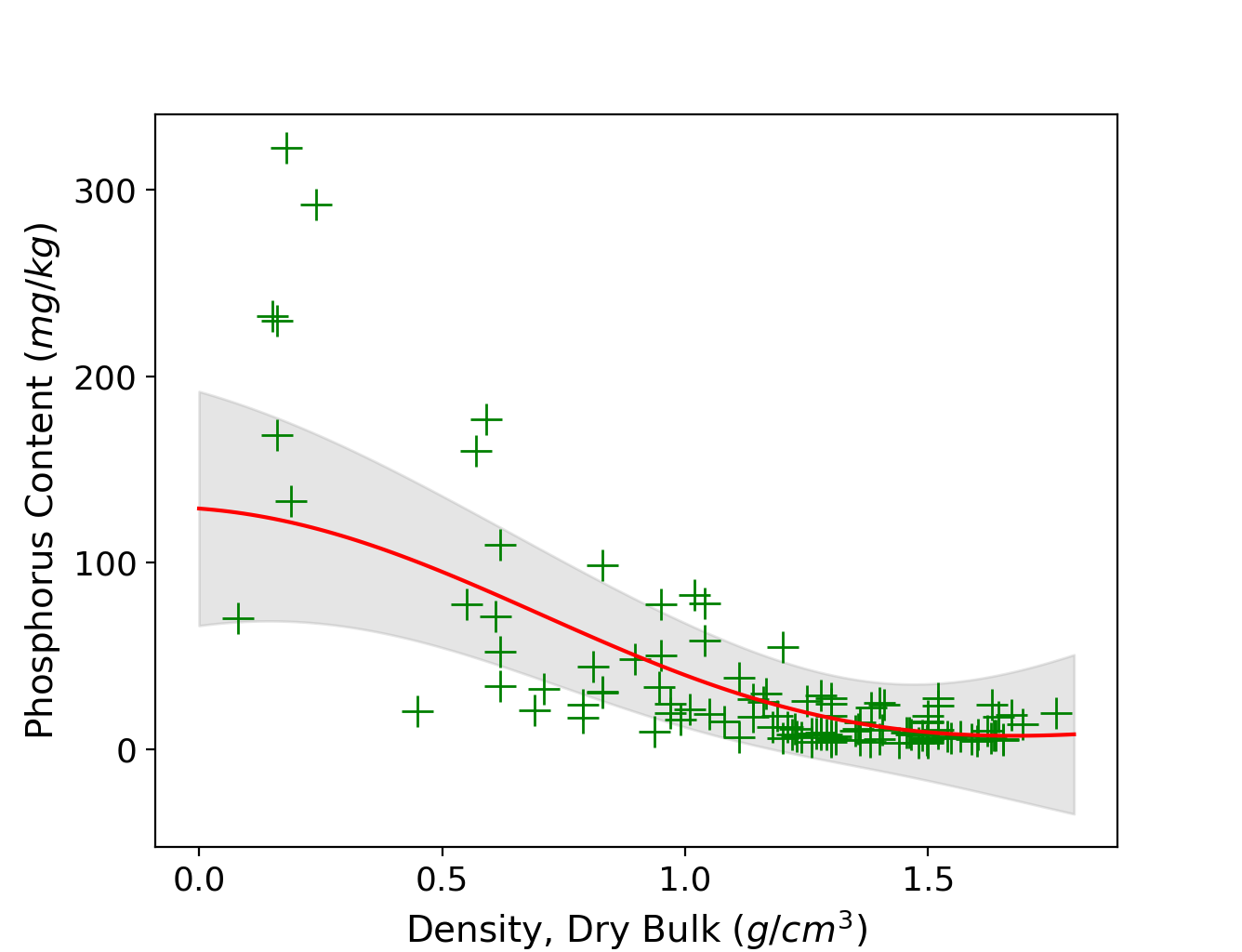}}
\subfigure[Heteroscedastic GP Fit]{\label{fig:het}\includegraphics[width=0.48\textwidth]{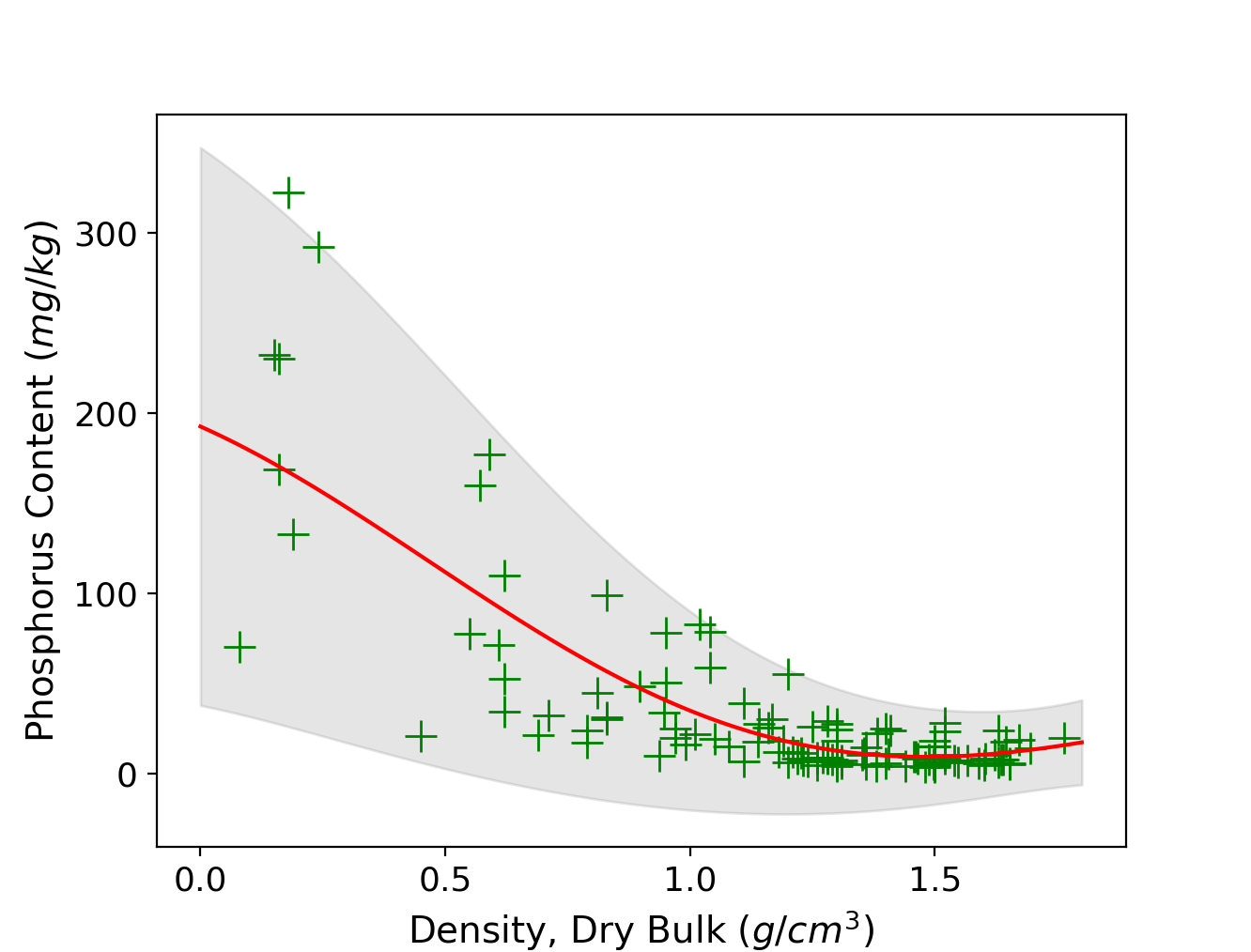}} 
\caption{Comparison of homoscedastic and heteroscedastic GP fits to the soil phosphorus fraction dataset \cite{houGlobalDatasetPlant2018}.}
\label{fig:soil}
\end{figure*}

In \autoref{fig:hist} we illustrate real-world sources of heteroscedasticity using the FreeSolv dataset of \cite{2017_Duarte}. The consequences of misrepresenting heteroscedastic noise as being homoscedastic, i.e. constant across the input domain, are illustrated using a second dataset \cite{houGlobalDatasetPlant2018} in \autoref{fig:soil}. The homoscedastic model can underestimate noise in certain regions of the input space which in turn could induce a Bayesian optimisation scheme to suggest values possessing large aleatoric noise. In an application such as high-throughput virtual screening \cite{2015_Knapp} the cost of misrepresenting noise during the screening process could lead to a substantial loss of time in material fabrication \cite{2017_Knapp}. In this paper we present a heteroscedastic Bayesian optimisation algorithm capable of both representing and minimising aleatoric noise in its suggestions.
Our contributions are:

\begin{enumerate}[label=(\arabic*)]
    \item The introduction of a novel combination of surrogate model and acquisition function designed to minimise heteroscedastic aleatoric uncertainty.
    \item A demonstration of our scheme's ability to outperform naive schemes based on homoscedastic Bayesian optimisation and random sampling on toy problems as well as two real-world scientific datasets.
    \item The provision of an open-source implementation.
    
\end{enumerate}

The paper is structured as follows: section 2 introduces related work on heteroscedastic Bayesian optimisation. Section 3 provides background on Bayesian optimisation and homoscedastic GP surrogate models. Section 4 provides background on the heteroscedastic GP surrogate model used in this work and introduces the novel HAEI and ANPEI acquisitions functions. Section 5 considers experiments on synthetic and scientific datasets possessing heteroscedastic noise where the goal is to be robust to, i.e. minimise, aleatoric noise in the suggestions. Section 6 presents an ablation study on noiseless tasks as well as tasks with homoscedastic and heteroscedastic noise in order to determine whether there is a detrimental effect to using a heteroscedastic surrogate when the noise properties of the problem are a priori unknown. Section 7 concludes with some limitations of the approach presented as well as fruitful sources for future work.

\section{Related Work}

The most similar work to our own is that of \cite{2017_Calandra} where experiments are reported on a heteroscedastic Branin-Hoo toy function using the variational heteroscedastic Gaussian process (GP) approach of \cite{2011_Lazaro}. This work defines and optimises a robustness index, making a compelling case for penalisation of aleatoric noise in real-world Bayesian optimisation problems. A modification to expected improvement (EI), expected risk improvement is introduced in \cite{2013_Kuindersma} and is applied to problems in robotics where robustness to aleatoric noise is desirable. In this framework however, the relative weights of performance and robustness cannot be tuned \cite{2017_Calandra}. \cite{2014_Assael, 2014_Ariizumi} implement heteroscedastic Bayesian optimisation but do not introduce an acquisition function that penalises aleatoric noise. \cite{2015_Sui, 2016_Berkenkamp} consider the related problem of safe Bayesian optimisation through implementing constraints in parameter space. In this instance, the goal of the algorithm is to enforce a performance threshold for each evaluation of the black-box function. Recently, the winners of the 2020 NeurIPS Black-Box Optimisation Competition applied non-linear output transformations in their solution to tackle heteroscedasticity. The authors however are not interested in explicitly penalising aleatoric noise in this case. In terms of acquisition functions, \cite{2009_Frazier, 2019_Letham} propose principled approaches to handling aleatoric noise in the homoscedastic setting that could be extended to the heteroscedastic setting. Our primary focus in this work however, is to highlight that heteroscedasticity in the surrogate model is beneficial and so an examination of a subset of acquisition functions is sufficient for this purpose. We take the opportunity here to note earlier unpublished workshop versions of this paper which consider the same problem \cite{neurips_sci, neurips_safety}.

\section{Background}

\subsection{Bayesian Optimisation}

Bayesian optimisation \cite{1964_Kushner, 1978_Mockus, 1998_Jones} solves the global optimisation problem defined as

\begin{align*}
\mathbf{x}^* = \operatorname*{arg\,min}_{\mathbf{x} \in \mathcal{X}} f(\mathbf{x})
\end{align*}

\noindent where $\mathbf{x}^*$ is the global optimiser of a black-box function $f: \mathcal{X} \rightarrow \mathcal{Y}$. $\mathcal{X}$ is the design space and is typically a compact subset of $\mathbb{R}^d$. What makes this optimisation problem practically relevant in applications are the following properties:

\begin{enumerate}
    \item Black-Box Objective: We do not have the analytic form of $f$. We can however evaluate $f$ pointwise anywhere in the design space $\mathcal{X}$.
    \item Expensive Evaluations: Choosing an input location $\mathbf{x}$ and evaluating $f(\mathbf{x})$ takes a very long time.
    
    \item Noise: The evaluation of a given $\mathbf{x}$ is a noisy process. In addition, this noise may vary across $\mathcal{X}$, making the underlying process heteroscedastic.
\end{enumerate}

We have a dataset $\boldsymbol{\mathcal{D}} = \{(\boldsymbol{x}_{i}, t_{i})\}^{n}_{i = 1}$ consisting of observations of the black-box function $f$ and fit a probabilistic surrogate model to these datapoints. We then leverage the predictive mean as well as the uncertainty estimates of the surrogate model to guide the acquisition of the next data point $\boldsymbol{x}_{n+1}$ according to a heuristic known as an acquisition function. In Bayesian optimisation, exact GPs are the most popular choice of surrogate model because of their ability to represent posterior uncertainty without resorting to approximate Bayesian inference.

\subsection{Gaussian Processes}

In the terminology of stochastic processes we may formally define a GP as follows:

\begin{defn} A GP \cite{2006_Rasmussen} is a collection of random variables, any finite number of which have a joint Gaussian distribution.\end{defn}

GPs can be used to set a prior over functions in Bayesian modelling applications. In this setting, the random variables consist of function values $f(\mathbf{x})$ at different locations $\mathbf{x}$ within the design space. The GP is characterised by a mean function

\begin{align*}
m(\mathbf{x}) = \mathbb{E}[f(\mathbf{x})]
\end{align*}

\noindent and a covariance function

\begin{align*}
k(\mathbf{x}, \mathbf{x'}) = \mathbb{E}[(f(\mathbf{x} - m(\mathbf{x}))(f(\mathbf{x'}) - m(\mathbf{x'}))].
\end{align*}

\noindent The process is written as follows

\begin{align*}
f(\mathbf{x}) \sim \mathcal{GP}\big(m(\mathbf{x}), k(\mathbf{x}, \mathbf{x'})\big).
\end{align*}

\noindent In our experiments, the prior mean function will be set to the empirical mean of the data. The covariance function or kernel computes the pairwise covariance between two random variables (function values). The covariance between a pair of output values $f(\mathbf{x})$ and $f(\mathbf{x'})$ is a function of an input pair $\mathbf{x}$ and $\mathbf{x'}$. As such, the kernel encodes smoothness assumptions about the latent function being modelled. The most widely-utilised kernel is the squared exponential (SE) kernel

\begin{equation}\label{eq:sq_exp_kernel}
    k_{\text{SQE}}(\boldsymbol{x}, \boldsymbol{x'}) = \sigma_{f}^{2}\cdot\text{exp}\Big(\frac{-\lVert\boldsymbol{x} - \boldsymbol{x'}\rVert^{2}}{2\ell^{2}}\Big)
\end{equation}

\noindent where $\sigma_{f}^2$ is the signal amplitude hyperparameter (vertical lengthscale) and $\ell$ is the (horizontal) lengthscale hyperparameter. Although \autoref{eq:sq_exp_kernel} is written with a single lengthscale shared across dimensions, for multidimensional input spaces we optimise a lengthscale per dimension. For consistency, we use the squared exponential kernel in all experiments reported in the main paper. In \ref{kernel_exps} we compare the performance of different kernels on a set of synthetic optimisation functions. For a more detailed introduction to GPs the reader is referred to \cite{2006_Rasmussen}.

\section{Heteroscedastic Bayesian Optimisation}
\label{method_section}

We wish to perform Bayesian optimisation whilst minimising input-dependent aleatoric noise. In order to represent input-dependent aleatoric noise, a heteroscedastic surrogate model is required. 

\subsection{The Most Likely Heteroscedastic Gaussian Process}

We adopt the most likely heteroscedastic Gaussian process (MLHGP) approach of \cite{2007_Kersting}, and for consistency, we use the same notation as the source work in our presentation. We have a dataset $\mathbf{\mathcal{D}} = \{(\mathbf{x}_{i}, t_{i})\}^{n}_{i = 1}$ in which the target values $t_{i}$ have been generated according to $t_{i} = f(\mathbf{x}_{i}) + \epsilon_{i}$. We assume independent Gaussian noise terms $\epsilon_{i} \sim \mathcal{N}(0, \sigma_{i}^2)$ with variances given by $\sigma_{i}^2 = r(\mathbf{x}_{i})$. In the heteroscedastic setting $r$ is typically a non-constant function over the input domain $\mathbf{x}$. In order to perform Bayesian optimisation, we wish to model the predictive distribution $P(\mathbf{t}^{*} \mid \mathbf{x}^{*}_{1}, \ldots, \mathbf{x}^{*}_{q})$ at the query points $\mathbf{x}^{*}_{1}, \ldots, \mathbf{x}^{*}_{q}$. Placing a GP prior on $f$ and taking $r(\textbf{x})$ as the assumed noise function, the predictive distribution is multivariate Gaussian $\mathcal{N}(\mathbf{\mu}^{*}, \Sigma^{*})$ with mean

\begin{equation*}
    \mathbf{\mu}^{*} = E[\mathbf{t^{*}}] = K^{*}(K + R)^{-1} \mathbf{t}
\end{equation*}

\noindent and covariance matrix

\begin{equation*}
    \Sigma^{*} = \text{var}[\mathbf{t^{*}}] = K^{**} + R^{*} - K^{*}(K + R)^{-1}K^{*T},
\end{equation*}

\noindent where $K \in \mathbb{R}^{n \times n}$, $K_{ij} = k(\mathbf{x}_{i},\mathbf{x}_{j})$, 
$K^{*} \in \mathbb{R}^{q \times n}$, $K_{ij}^{*} = k(\mathbf{x}_{i}^{*},\mathbf{x}_{j})$, $K^{**} \in \mathbb{R}^{q \times q}$, $K^{**}_{ij} = k(\mathbf{x}^{*}_{i}, \mathbf{x}^{*}_{j})$, $\mathbf{t} = (t_{1}, t_{2}, \ldots, t_{n})^{T}$, $R = \text{diag}(\mathbf{r}) \text{ with } \mathbf{r} = (r(\mathbf{x}_{1}),r(\mathbf{x}_{2}), \ldots, r(\mathbf{x}_{n}))^{T}$, and $R^{*} = \text{diag}(\mathbf{r}^{*}) \text{ with }\mathbf{r}^{*} = (r(\mathbf{x}^{*}_{1}), r(\mathbf{x}^{*}_{2}),\ldots, r(\mathbf{x}^{*}_{q}))^{T}$.\\

\noindent The MLHGP algorithm \cite{2007_Kersting} executes the following steps:

\begin{enumerate}
    \item Estimate a homoscedastic GP, $G_1$ on the dataset $\mathbf{\mathcal{D}} = \{(\mathbf{x}_{i}, t_{i})\}^{n}_{i = 1}$
    \item Given $G_1$, we estimate the empirical noise levels for the training data using $z_i = \log(\text{var}[t_i, G_1(\textbf{x}_i, \mathcal{D})])$ where $\text{var}[t_i, G_1(\textbf{x}_i, \mathcal{D})] \approx \frac{1}{s} \: \sum_j^s \: 0.5 \: (t_i - t_i^j)^2$ with $t_i^j$ a sample from the predictive distribution induced by the GP at $\mathbf{x}_i$, forming a new dataset 
    $\mathbf{\mathcal{D}'} = \{(\mathbf{x}_{i}, z_{i})\}^{n}_{i = 1}$
    \item Estimate a second GP, $G_2$ on $\mathbf{\mathcal{D}'}$.
    \item Estimate a combined GP, $G_3$ on $\mathbf{\mathcal{D}}$ using $G_2$ to predict the logarithmic noise levels $r_i$.
    \item If not converged, set $G_3$ to $G_1$ and repeat.
\end{enumerate}

\noindent In essence, the defining characteristic of the MLHGP approach is that $G_1$ learns the latent function and $G_2$ learns the noise function. 

\subsection{Bayesian Optimisation with Aleatoric Noise Penalisation}

Our heteroscedastic Bayesian optimisation problem may be framed as

\begin{equation*}
    \boldsymbol{x}^{*} = \operatorname*{arg\,min}_{\boldsymbol{x} \in \chi} h(\boldsymbol{x}),
\end{equation*}

\noindent where the black-box objective $h$, to be minimised has the form

\begin{equation*}
    \label{noise_equation}
    h(\boldsymbol{x}) = \alpha f(\boldsymbol{x}) + (1 - \alpha)g(\boldsymbol{x}).
\end{equation*}

\noindent where $f(\boldsymbol{x})$ is the black-box function of the principal objective i.e. the objective corresponding to classical Bayesian optimisation where noise is not optimised, and $g(\boldsymbol{x})$ is the latent heteroscedastic noise function which governs the magnitude of the noise at a given input location $\boldsymbol{x}$. $\alpha$ is a parameter chosen, for the purposes of evaluation, by a domain expert that trades off the weight of the principal objective relative to the noise objective. It is worth noting that $\alpha$ is a parameter that affects only the evaluation of an algorithm and not the execution. The evaluation criteria however, will dictate the optimal hyperparameters of the acquisition function.

\subsection{Heteroscedastic Acquisition Functions}

We investigate extensions of the expected improvement \cite{1998_Jones} acquisition criterion, the form of which may be written in terms of the targets $t$ and the incumbent best objective function value, $\eta$, found so far as

\begin{equation*}
    \text{EI}(\boldsymbol{x}) = \mathbb{E}\big[\,(\eta - t)_{+}\big] = \int_{-\infty}^{\infty}(\eta - t)_{+}\,p(t\,|\,\boldsymbol{x})\,dt
\end{equation*}

\noindent where $p(t\,|\,\boldsymbol{x})$ is the posterior predictive marginal density of the objective function evaluated at $\boldsymbol{x}$. \hspace{3mm}$(\eta - t)_{+} \equiv \text{max}\,(0,\, \eta - t)$ is the improvement over the incumbent best objective function value $\eta$. Evaluations of the objective are noisy in all of the problems we consider and so we use expected improvement with plug-in \cite{2013_Picheny}, the plug-in value being the GP predictive mean \cite{2008_Vasquez}.\\

\noindent We propose two extensions to the expected improvement criterion. The first is an extension of the augmented expected improvement criterion

\begin{equation*}
    \text{AEI}(\boldsymbol{x}) = \mathbb{E}\big[(\eta - t)_{+}\big] \Bigg( 1 - \frac{\sigma_n}{\hphantom{e}\vphantom{\bigg|}\sqrt{\text{var}[t] + \sigma_n^{2}}\hphantom{e}} \Bigg),
\end{equation*}

\noindent of \cite{2006_Huang} where $\sigma_n$ is the fixed aleatoric noise level. AEI is introduced as a heuristic for the optimisation of noisy functions. EI is recovered in the case that $\sigma_n^{2} = 0$ and in the case that $\sigma_n^2 > 0$ AEI operates as a rescaling of the EI acquisition function, penalising test locations where the GP predictive variance is small relative to the fixed noise level $\sigma_n^2$. We extend AEI to the heteroscedastic setting by exchanging the fixed aleatoric noise level with the input-dependent one:

\begin{equation}
\label{eq:12}
    \text{HAEI}(\boldsymbol{x}) = \mathbb{E}\big[(\eta - t)_{+}\big] \Bigg( 1 - \frac{\gamma\sqrt{r(\boldsymbol{x})}}{\hphantom{e}\vphantom{\big|}\sqrt{\text{var}[t] + \gamma^2r(\boldsymbol{x})}\hphantom{e}} \Bigg), 
\end{equation}

\noindent where $r(\boldsymbol{x})$ is the predicted aleatoric uncertainty at input $\boldsymbol{x}$ under the MLHGP and $\text{var}[t]$ is the predictive variance of the MLHGP at input $\boldsymbol{x}$. $\gamma$ in this instance is defined to be a positive penalty parameter for regions with high aleatoric noise. \\

\begin{proposition}[Limit of Large Epistemic Uncertainty]
\label{prop:prop1}

The HAEI acquisition function reduces to EI when the ratio of epistemic uncertainty to aleatoric uncertainty is much greater than $\gamma^2$. \end{proposition}

\begin{proof}

Let $k = \frac{\text{var}[t]}{r(\boldsymbol{x})}$ denote the ratio of epistemic to aleatoric uncertainty at an arbitrary input location $\boldsymbol{x}$. Dividing the numerator and the denominator of the second term in the second factor of \autoref{eq:12} by $\sqrt{r(\boldsymbol{x})}$ yields

\begin{equation}
\label{eq:13}
    \text{HAEI}(\boldsymbol{x}) = \text{EI}(\boldsymbol{x})\Bigg( 1 - \frac{\gamma}{\hphantom{e}\vphantom{\big|}\sqrt{k + \gamma^2}\hphantom{e}} \Bigg).
\end{equation}

\noindent Taking the limit analytically as $k$ tends to infinity and assuming finite $\gamma$

\begin{equation*}
    \lim_{k \to \infty}\text{EI}(\boldsymbol{x})\Bigg( 1 - \frac{\gamma}{\hphantom{e}\vphantom{\big|}\sqrt{k + \gamma^2}\hphantom{e}} \Bigg) = \text{EI}(\boldsymbol{x}),
\end{equation*}

\noindent recovers the expected improvement acquisition. 

\end{proof}

\begin{proposition}[Limit of Large Aleatoric Uncertainty]\label{prop2}

The HAEI acquisition function goes to zero as the ratio of epistemic uncertainty to aleatoric uncertainty goes to zero. \end{proposition}

\begin{proof}

Taking the limit as $k$ tends to zero in \autoref{eq:13} yields 

\begin{equation*}
    \lim_{k \to 0}\text{EI}(\boldsymbol{x})\Bigg( 1 - \frac{\gamma}{\hphantom{e}\vphantom{\big|}\sqrt{k + \gamma^2}\hphantom{e}} \Bigg) = 0.
\end{equation*}

\end{proof}

\begin{remark} In the limit of large aleatoric uncertainty there is an approximation that is linear in $k$ for the HAEI scaling factor.
\end{remark}

Letting $S(k) = 1 - \frac{\gamma}{\sqrt{k + \gamma^2}}$ such that $\text{HAEI} = \text{EI}(\boldsymbol{x})S(k)$, consider the MacLaurin expansion of $S(k)$,

\begin{equation*}
    S(k) = S(0) + S'(0)k + \frac{S''(0)}{2!}k^2 + \frac{S'''(0)}{3!}k^3 + \dotsc ,
\end{equation*}

\noindent Dropping terms of $O(k^2)$ and higher we obtain

\begin{equation*}
    S(k) \approx \frac{k}{2\gamma^2}.
\end{equation*}

\begin{figure*}[t]
\centering
    \includegraphics[width=.7\textwidth]{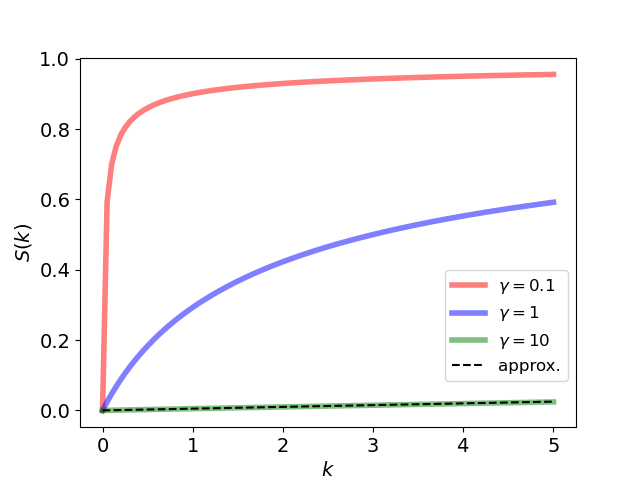}
    \caption{The HAEI scaling factor $S(k)$, now written as a function of $k$ for different values of $\gamma$. When $k$, the ratio of epistemic to aleatoric uncertainty is small, the scaling factor goes to zero to reflect the penalty for regions of high aleatoric uncertainty. $\gamma$ controls the decay rate of this penalty. Also shown is the linear approximation to the scaling factor for $\gamma = 10$.}
    \label{scalig}
\end{figure*}

\noindent This approximation may be used when $k$ is small relative to $\gamma$ and could provide guidance in setting the $\gamma$ parameter if prior knowledge about $k$ and the desired trade-off between the principal and noise objectives is available. In \autoref{scalig} we provide insight into the effect that different values of $\gamma$ will have on the scaling factor $S(k)$. \\

In addition to HAEI, we propose a simple modification to EI that explicitly penalises regions of the input space with large aleatoric noise. We call this acquisition function aleatoric noise-penalised expected improvement (ANPEI) and denote it

\begin{equation}\label{eq:anpei_equation}
    \text{ANPEI} = \beta\text{EI}(\boldsymbol{x}) - (1 - \beta)\sqrt{r(\boldsymbol{x})},
\end{equation}

\noindent where $\beta$ is a scalarisation constant. In the multiobjective optimisation setting a particular value of $\beta$ will correspond to a point on the Pareto frontier. We showcase the advantages of both HAEI and ANPEI acquisition functions in conjunction with the MLHGP surrogate model in \autoref{experiments}.


\section{Experiments on Robustness to Aleatoric Uncertainty}
\label{experiments}

\begin{figure*}
\centering
\subfigure[Latent Function]{\label{fig:sin_1_first}\includegraphics[width=0.32\textwidth]{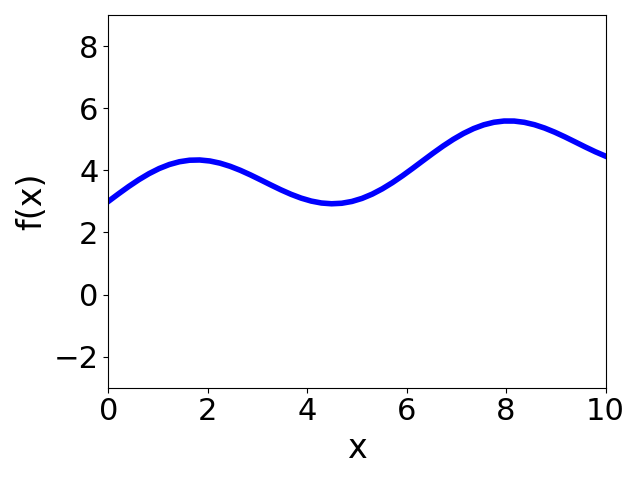}}
\subfigure[Noise Function ]{\label{fig:sin_2_first}\includegraphics[width=0.32\textwidth]{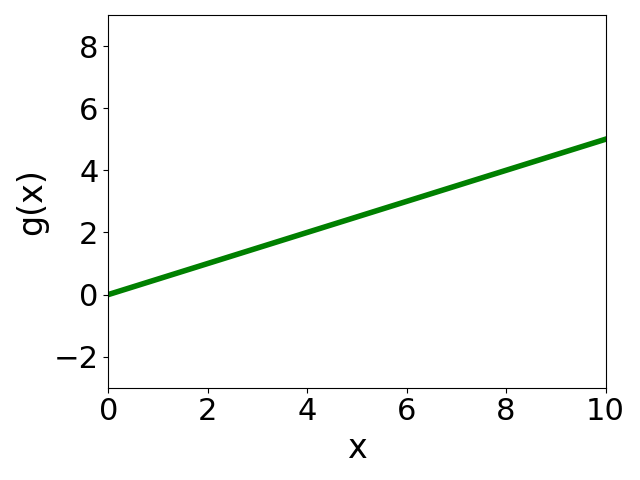}}
\subfigure[Objective Function]{\label{fig:sin_3_first}\includegraphics[width=0.32\textwidth]{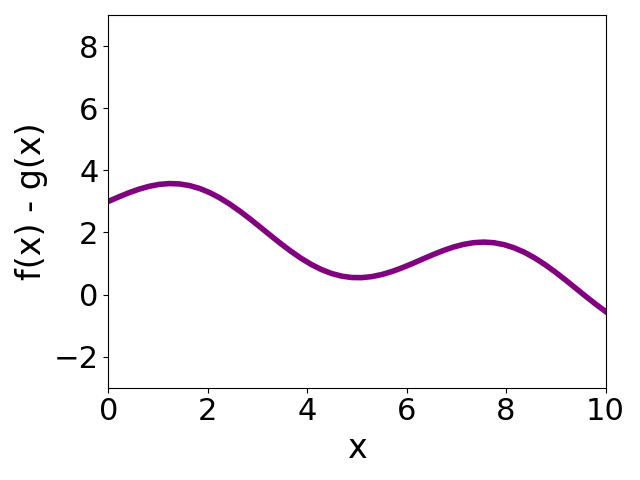}} 
\caption{Illustrative Toy Problem. The latent function in a) is corrupted with heteroscedastic Gaussian noise according to the function in b) where $g(x)$ is a constant multiplier of a sample from a standard Gaussian. The combined objective is given in c) and is obtained by subtracting the noise function from the latent function.}
\label{fig:sin_first}
\end{figure*}

\begin{figure*}[t]
\centering
    \includegraphics[width=.5\textwidth]{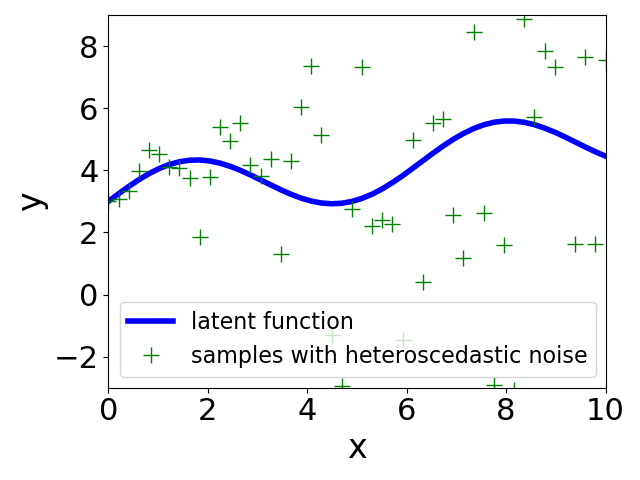}
    \caption{Noisy samples $y_i = f(x_i) + g(x_i)\epsilon$ from the heteroscedastic sin wave function.}
    \label{sin_wave_samples}
\end{figure*}

\subsection{Implementation}


Experiments were run using a custom NumPy \cite{2020_numpy} implementation of GP regression and MLHGP regression. All code to reproduce the experiments is available at \url{https://github.com/Ryan-Rhys/Heteroscedastic-BO}. The squared exponential kernel was chosen as the covariance function for both the homoscedastic GP as well as $G_1$ and $G_2$ of the MLHGP. Across all datasets, the lengthscales, $\ell$, of the homoscedastic GP were initialised to 1.0 for each input dimension. The signal amplitude $\sigma_{f}^{2}$ was initialised to a value of 1.0. The lengthscale, $\ell$, of $G_2$ of the MLHGP \cite{2007_Kersting} was initialised to 1.0, the initial noise level of $G_2$ was set to 1.0. The EM-like procedure required to train the MLHGP was run for 10 iterations and the sample size required to construct the variance estimator producing the auxiliary dataset was 100. All standard error confidence bands are computed using 50 independent random seed initialisations. Hyperparameter values, including the noise level of the homoscedastic GP, were obtained by optimising the marginal likelihood using the scipy implementation of the L-BFGS-B optimiser \cite{1997_Zhu}, taking the best of 20 random restarts. The objective function is

\begin{equation*}
    \label{example1}
    h(x) = \alpha f(x) - (1 - \alpha)g(x)
\end{equation*}

\noindent for the one-dimensional sin wave experiment which is a maximisation problem and as such has a subtractive penalty for regions of large noise. For the remaining experiments, which are minimisation problems, the objective is

\begin{equation}
    \label{example2}
    h(\boldsymbol{x}) = \alpha f(\boldsymbol{x}) + (1 - \alpha)g(\boldsymbol{x})
\end{equation}

\noindent The sin wave and Branin-Hoo tasks are initialised with 25 and 100 data points respectively drawn uniformly at random within the bounds of the design space. The soil and FreeSolv experiments are initialised with 36 and 129 data points respectively drawn uniformly at random from the datasets. $\alpha$ is set to 0.5 for all experiments while $\beta$ is set to 0.5, $\frac{1}{11}$, 0.5 and 0.5 for the sin, Branin-Hoo, soil and FreeSolv experiments. $\gamma$ is set to 1, 500, 1 and 1 for the sin, Branin-Hoo, soil and FreeSolv experiments. We run 5 acquisition functions in all experiments: random sampling, homoscedastic EI, AEI, HAEI and ANPEI. Homoscedastic EI is included as a baseline to demonstrate the difference consideration of aleatoric noise yields in the optimisation of the objective. AEI is included to demonstrate the difference consideration of heteroscedastic aleatoric noise yields and random sampling is included as a baseline as it is known to be highly competitive with Bayesian optimisation in noisy settings.

\subsection{Heteroscedastic Sin Wave Function}

\begin{figure*}
\centering
\subfigure[Best Objective Value Found so Far]{\label{fig:bo_1}\includegraphics[width=0.486\textwidth]{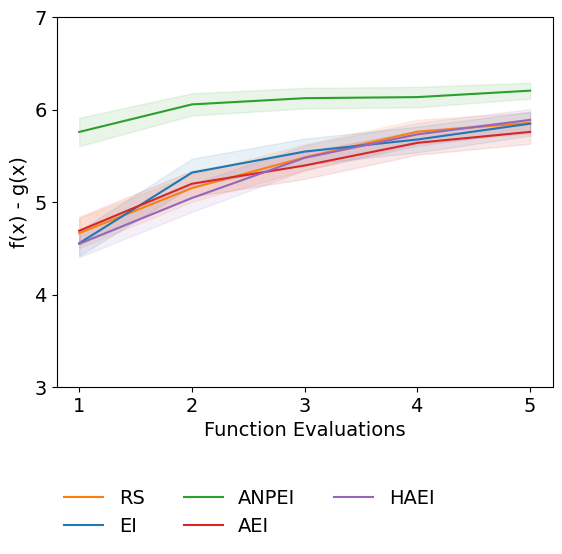}}
\subfigure[Lowest Aleatoric Noise Found so Far ]{\label{fig:bo_2}\includegraphics[width=0.501\textwidth]{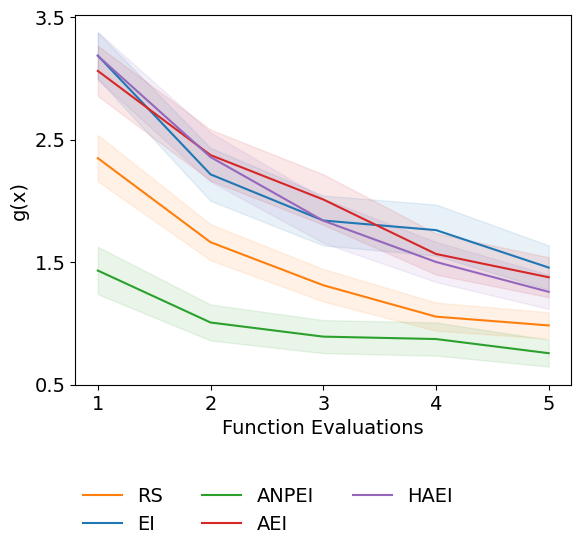}}
\caption{Comparison of heteroscedastic and homoscedastic Bayesian optimisation on the sin wave problem. (a) shows the optimisation of $h(x) = f(x) - g(x)$ (higher is better) whereas (b) shows the values $g(x)$ obtained over the course of the optimisation of $h(x)$. This latter plot demonstrates the propensity of ANPEI to seek low aleatoric noise solutions.}
\label{fig:sin_bayesopt}
\end{figure*}

The objective function has the form 

\begin{equation*}
    \label{example3}
    h(x) = f(x) - g(x)
\end{equation*}

\noindent where $f(x) = \sin(x) + 0.2(x) + 3$ and $g(x) = 0.5(x)$. In this instance $\alpha$ from \autoref{example1} has a setting of $0.5$ but we omit it explicitly as the objectives have equal weight. Over the course of the experiment samples 

\begin{align*}
    y_i = f(x_i) + g(x_i)\epsilon, \: \: \: \: \: \:\epsilon \sim \mathcal{N}(0, 1)
\end{align*}

\noindent are observed. The problem setup is depicted in \autoref{fig:sin_first} and \autoref{sin_wave_samples}. The Bayesian optimisation problem is constructed such that the first maximum in \autoref{fig:sin_1_first} is to be preferred as samples from this region of the input space will have low aleatoric noise. The black-box objective in \autoref{fig:sin_3_first} illustrates this trade-off. In \autoref{fig:sin_bayesopt} we compare the performance of all surrogate model/acquisition function combinations. We observe the low aleatoric noise-seeking behaviour of HAEI and ANPEI on $g(x)$ as well as their ability to optimise the composite objective $h(x)$.

\subsection{Heteroscedastic Branin-Hoo Function}\label{het_bran_section}

In the second experiment we consider the objective

\begin{align*}
    h(\boldsymbol{x}) = f(\boldsymbol{x}) + g(\boldsymbol{x})
\end{align*}

\noindent with an additive penalty because the task is a minimisation problem and an $\alpha$ setting of $0.5$ for equal-weight objectives.

\begin{align}
\label{branin_eq}
f(\boldsymbol{x})=\frac{1}{51.95}\left[\left(\bar{x}_{2}-\frac{5.1 \bar{x}_{1}^{2}}{4 \pi^{2}}+\frac{5 \bar{x}_{1}}{\pi}-6\right)^{2}+\left(10-\frac{10}{8 \pi}\right) \cos \left(\bar{x}_{1}\right)-44.81\right]
\end{align}

\noindent with $\bar{x}_{1} = 15x_1 - 5$, $\bar{x}_{2} = 15x_2$ and $\boldsymbol{x} = (x_1, x_2)$ is the standardised Branin-Hoo function introduced in \cite{2013_Picheny}. The noise function $g(\boldsymbol{x})$ is in this instance



\begin{align}
\label{branin_noise_eq}
    g(\boldsymbol{x}) = 15 - 8x_{1} + 8x_{2}^2.
\end{align}

\noindent Samples are again generated according to

\begin{align*}
    y_i = f(\boldsymbol{x}_i) + g(\boldsymbol{x}_i)\epsilon, \: \: \: \: \: \:\epsilon \sim \mathcal{N}(0, 1)
\end{align*}

\begin{figure*}
\centering
\subfigure[Latent Function]{\label{fig:branin_1}\includegraphics[width=0.32\textwidth]{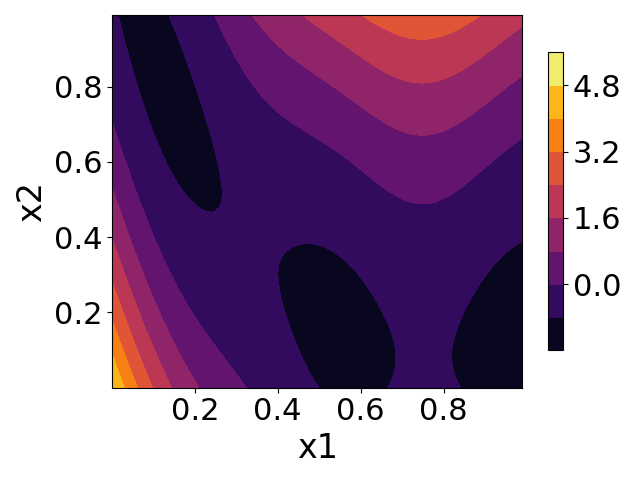}}
\subfigure[Non-linear Noise Function]{\label{fig:branin_2}\includegraphics[width=0.32\textwidth]{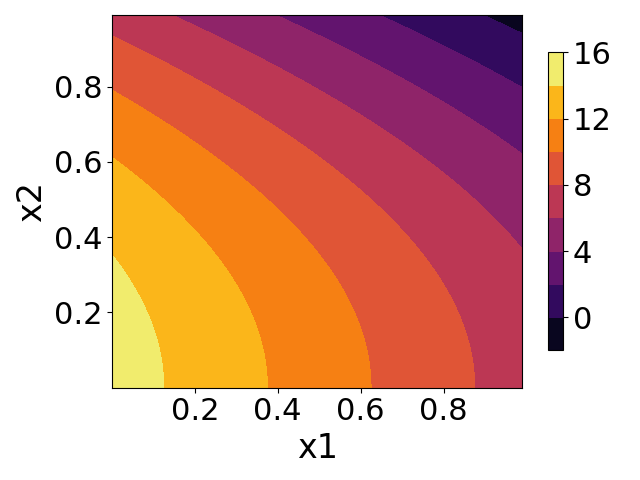}}
\subfigure[Objective Function]{\label{fig:branin_3}\includegraphics[width=0.32\textwidth]{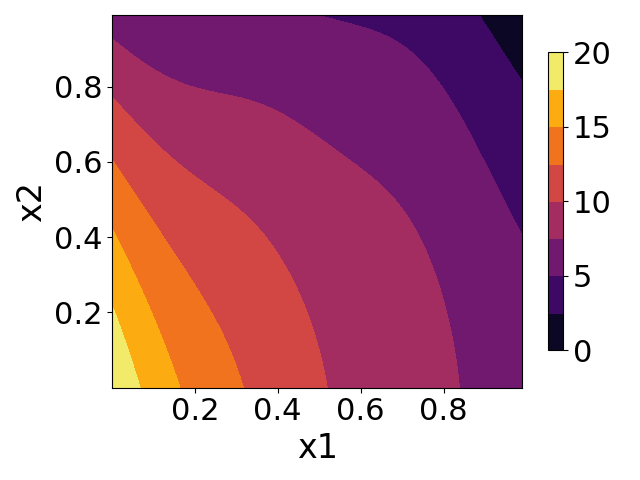}} 
\caption{Branin-Hoo Optimisation Problem. The latent function in a) is corrupted by heteroscedastic Gaussian noise function according to the function in b) The combined objective function is given in c) and is obtained by summing the functions in a) and b). The sum is required to penalise regions of large aleatoric noise because the objective is being minimised.}
\label{fig:branin}
\end{figure*}

\begin{figure*}
\centering
\subfigure[Best Objective Value Found so Far]{\label{fig:bo_1_branin}\includegraphics[width=0.49\textwidth]{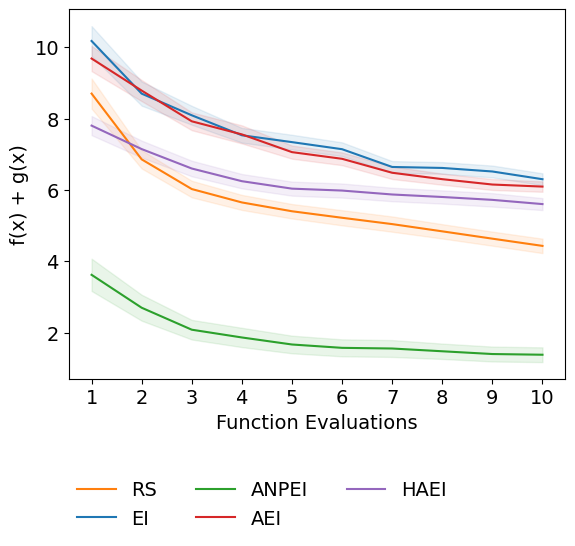}}
\subfigure[Lowest Aleatoric Noise Found so Far ]{\label{fig:bo_2_branin}\includegraphics[width=0.49\textwidth]{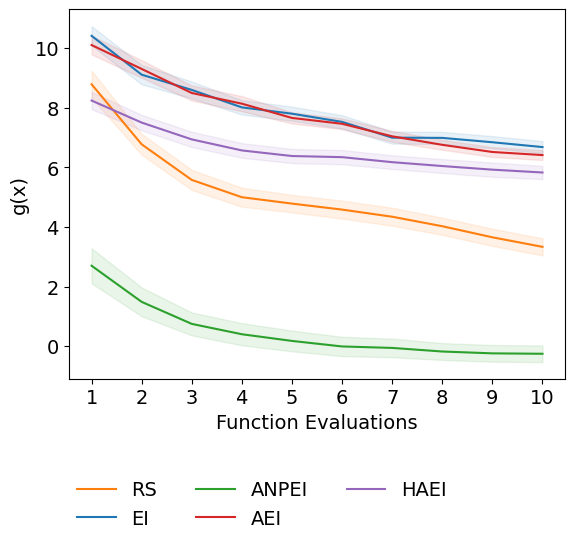}}
\caption{Comparison of heteroscedastic and homoscedastic Bayesian optimisation on the Branin-Hoo problem. (a) shows the optimisation of $h(\boldsymbol{x}) = f(\boldsymbol{x}) + g(\boldsymbol{x})$ (lower is better) whereas (b) shows the values $g(\boldsymbol{x})$ obtained over the course of the optimisation of $h(\boldsymbol{x})$.}
\label{fig:branin_bayesopt}
\end{figure*}

The problem setup is shown in \autoref{fig:branin} and the performance of all surrogate model/acquisition function pairs is depicted in \autoref{fig:branin_bayesopt}. The gulf in performance between the heteroscedastic and homoscedastic surrogate models is more pronounced in this case because the noise function is more severe relative to the sin wave problem.
 
\subsection{Soil Phosphorus Fraction Optimisation}

In this experiment we consider the optimisation of the phosphorus fraction of soil. Soil phosphorus is an essential nutrient for plant growth and is widely used as a fertiliser in agriculture. While the amount of arable land worldwide is declining, global population is expanding concomitantly with food demand. As such, understanding the availability of plant nutrients that increase crop yield is a topic worthy of attention. To this end, \cite{houGlobalDatasetPlant2018} have curated a dataset on soil phosphorus, relating phosphorus content to variables such as soil particle size, total nitrogen, organic carbon and bulk density. We choose to study the relationship between bulk soil density and the phosphorus fraction, the goal being to minimise the phosphorus content of soil subject to heteroscedastic noise. In lieu of performing a formal test for heteroscedasticity, we provide evidence that there is heteroscedasticity in the dataset by comparing the fits of a homoscedastic GP and the MLHGP in \autoref{fig:soil} and provide a predictive performance comparison based on negative log predictive density values in Appendix A. 

In this problem, we do not have access to a continuous-valued black-box function or a ground truth noise function. As such, the surrogate models were initialised with a subset of the data and the query locations selected by Bayesian optimisation were mapped to the closest datapoints in the heldout data. The following kernel smoothing procedure was used to generate pseudo ground-truth noise values:

\begin{enumerate}[label=(\arabic*)]
    \item Fit a homoscedastic GP to the full dataset.
    \item At each point $x_i$, compute the corresponding squared error $s_i^2 = (y_i - \mu(x_i))^2$.
    \item Estimate variances by computing a moving average of the squared errors, where the relative weight of each $s_i^2$ was assigned with a Gaussian kernel.
\end{enumerate}

\begin{figure*}
\centering
\subfigure[Best Objective Value Found so Far]{\label{fig:bo_1_soil}\includegraphics[width=0.50\textwidth]{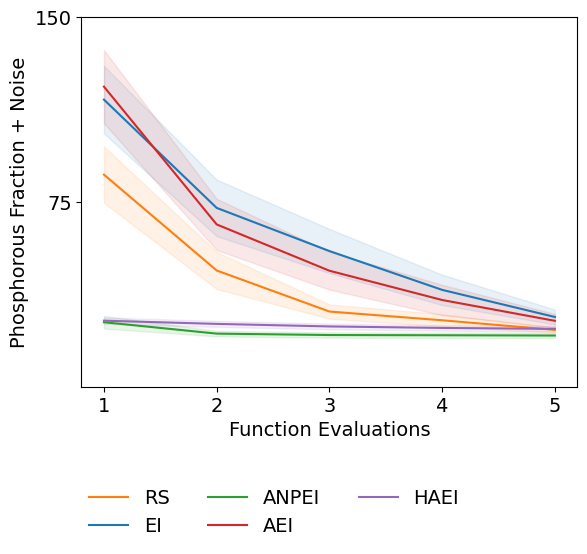}}
\subfigure[Lowest Aleatoric Noise Found so Far ]{\label{fig:bo_2_soil}\includegraphics[width=0.488\textwidth]{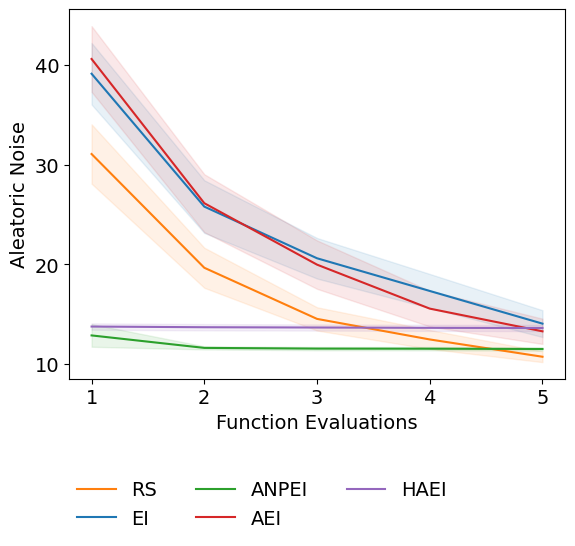}}
\caption{Comparison of heteroscedastic and homoscedastic Bayesian optimisation on the soil phosphorus fraction optimisation problem. (a) shows the optimisation of $h(x) = f(x) + g(x)$ (lower is better) where $x$ is the dry bulk density of the soil. (b) shows the values $g(x)$ obtained over the course of the optimisation of $h(x)$.}
\label{fig:soil_bayesopt}
\end{figure*}

\noindent The performances of heteroscedastic and homoscedastic Bayesian optimisation are compared in \autoref{fig:soil_bayesopt}. Given that regions of low phosphorus fraction coincide with regions of small aleatoric noise, we apply an $\alpha$ value of $\frac{1}{6}$ to the composite objective $h(x)$ to admit a finer granularity for distinguishing between degrees of low aleatoric noise in the solutions.

\subsection{Molecular Hydration Free Energy Optimisation}

\begin{figure*}
\centering
\subfigure[Best Objective Value Found so Far]{\label{fig:bo_freesolv}\includegraphics[width=0.498\textwidth]{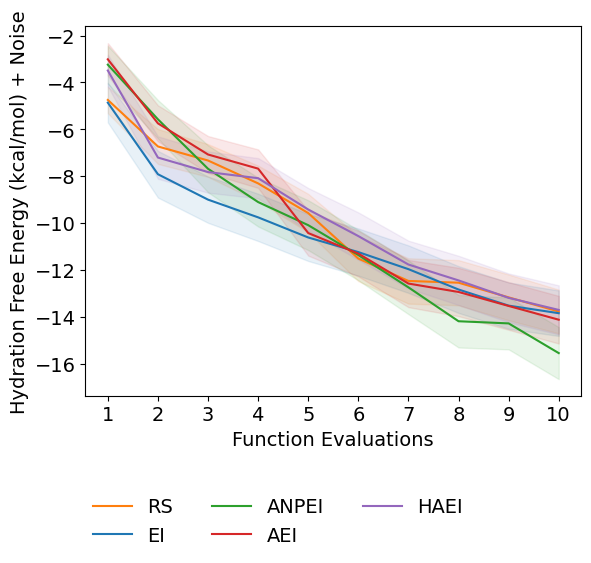}}
\subfigure[Lowest Aleatoric Noise Found so Far ]{\label{fig:bo_2_freesolv}\includegraphics[width=0.49\textwidth]{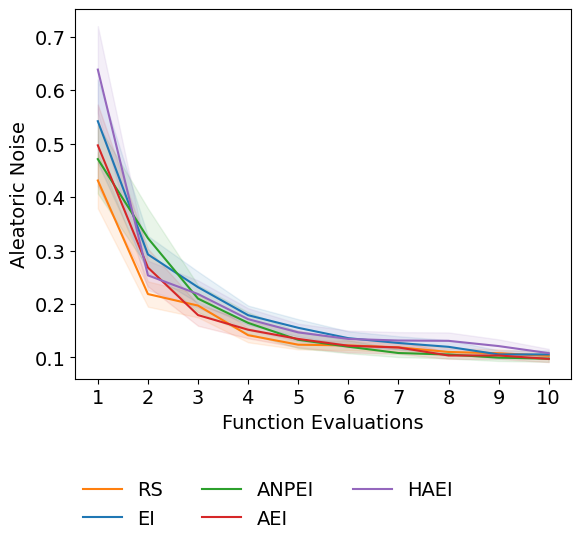}}
\caption{Comparison of heteroscedastic and homoscedastic Bayesian optimisation on the FreeSolv hydration free energy optimisation problem. (a) shows the optimisation of $h(\boldsymbol{x}) = f(\boldsymbol{x}) + g(\boldsymbol{x})$ (lower is better) where $\boldsymbol{x}$ is the fragment set of molecular descriptors, $f(\boldsymbol{x})$ is the hydration free energy and $g(\boldsymbol{x})$ is the aleatoric noise. (b) shows the values $g(\boldsymbol{x})$ obtained over the course of the optimisation of $h(\boldsymbol{x})$.}
\label{fig:freesolv_bayesopt}
\end{figure*}

We perform a retrospective virtual screening experiment with the aim of identifying molecules with favourable hydration free energy, a property important in determining the binding affinity of a drug candidate. Experiments were performed with an initialisation of 129 out of the 642 molecules in the FreeSolv dataset \cite{2014_Mobley, 2017_Duarte} over 10 iterations of data collection. Unlike the soil phosphorus fraction dataset, ground truth measurement error (aleatoric noise $g(\mathbf{x})$) values are available for the FreeSolv dataset. The remaining 513 molecules were reserved as a heldout set where at each iteration of data collection one of the heldout molecules was selected. Chemical fragments computed using RDKit \cite{rdkit} were used as the molecular representation based on the fact that these global features, unlike local Morgan fingerprints, act as good predictors of the hydration free energy. The fragment features were projected down to 14 components using principal component analysis, retaining more than 90\% of the variance on average across random trials. The results are shown in \autoref{fig:freesolv_bayesopt}. Compared to previous experiments, the noise is smaller in this instance relative to the magnitude of the hydration free energy (Signal-to-noise ratio of approximately 10) and as such the heteroscedastic modelling problem is more difficult, leading to only very marginal gains in obtaining low noise solutions. While ANPEI obtains the lowest objective function value over the Bayesian optimisation trace, the results are unlikely to be statistically significant according to the standard error bands.

\subsection{Heteroscedastic Acquisition Function Hyperparameters}

\begin{figure*}
\centering
\subfigure[ANPEI]{\label{fig:bo_branin_hyper}\includegraphics[width=0.49\textwidth]{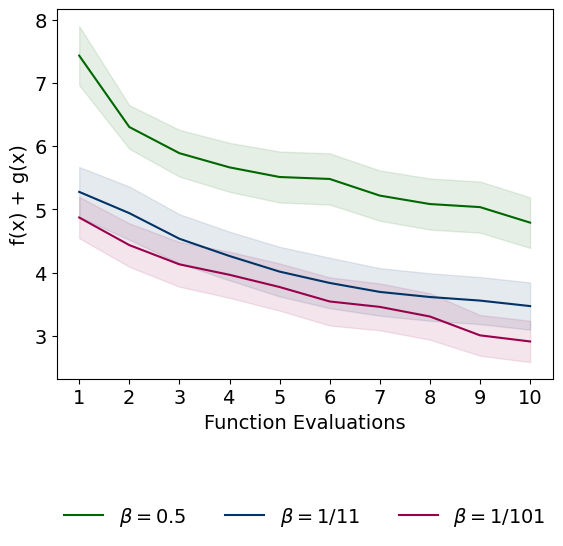}}
\subfigure[HAEI]{\label{fig:bo_2_branin_hyper}\includegraphics[width=0.49\textwidth]{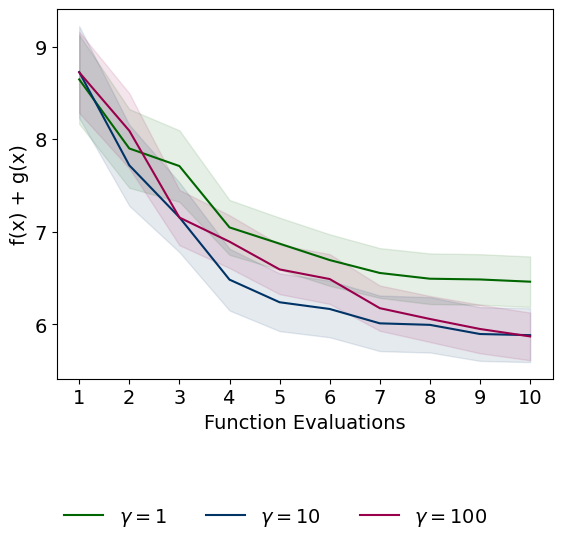}}
\caption{Performance of ANPEI and HAEI plotted for different values of the $\beta$ and $\gamma$ hyperparameters respectively. \textit{Smaller} values of $\beta$ encourage avoidance of regions of high aleatoric noise whilst \textit{larger} values of $\gamma$ encourage avoidance of regions of high aleatoric noise.}
\label{fig:weight_comparison}
\end{figure*}

The $\beta$ hyperparameter of ANPEI in \autoref{eq:anpei_equation} and the $\gamma$ hyperparameter of HAEI in \autoref{eq:13} are designed to modulate the avoidance of aleatoric noise in the acquisitions. In \autoref{fig:weight_comparison} we offer some intuition as to the effect of various settings of $\beta$ and $\gamma$ by examining the heteroscedastic Branin-Hoo function introduced in \autoref{het_bran_section}. The results demonstrate that the performance of the algorithms is strongly dependent on the setting of the $\beta$ hyperparameter for ANPEI whereas $\gamma$ is less influential on the performance of HAEI. It is worth noting in \autoref{fig:bo_2_branin_hyper} that if too large a value of $\gamma$ is chosen the principal objective $f(\mathbf{x})$ may be compromised through overly aggressive avoidance of aleatoric noise. In practice choosing the value of $\beta$ in line with the value of the evaluation criterion parameter $\alpha$ in \autoref{example2} is likely to be a sensible approach i.e. if the noise objective is more important relative to the principal objective by a factor of 10 then the value of $\beta$ should be $\frac{1}{11}$.

\subsection{Robustness Experiments Summary}


The experiments of this section provide strong evidence that modelling heteroscedasticity in Bayesian optimisation is a useful approach for problems in which there is a strong degree of aleatoric noise present. The ANPEI acquisition tends to outperform HAEI on the majority of the tasks where there is a small degree of aleatoric noise whilst the acquisitions are more evenly matched when the extent of the aleatoric noise is high. The outstanding questions for these methods however, is how well they perform on tasks where heteroscedastic noise is not present. Such a situation may easily arise for real-world problems where the noise properties of the tasks are a prior unknown and as such, it is important to ascertain whether there is a deleterious effect on performance in noiseless and homoscedastic noise settings.

\section{Ablation Study on Noiseless, Homoscedastic Noise and Heteroscedastic Noise Tasks}\label{first_ablation}

In this section we perform an ablation study where components of the ablation constitute different noise properties. We examine the noiseless case as a base task before adding first a homoscedastic noise component and second, a heteroscedastic noise component. Additionally, we examine the effect of the size of the initialisation grid on performance in the heteroscedastic noise tasks.

\subsection{Ablation}

The ablation study makes use of three synthetic optimisation functions: The Branin-Hoo function, the Hosaki function and the Goldstein-Price function. The form of the Branin-Hoo function is the same standardised Branin-Hoo function introduced in \autoref{branin_eq} with heteroscedastic noise function given in \autoref{branin_noise_eq}. The Hosaki function, defined on the domain $x_1, x_2 \in [0, 5]$, is


\begin{align*}
\text{Hosaki}(x_1, x_2) = \Big(1 - 8x_1 + 7{x_1}^2 - \frac{7}{3} {x_1}^3 + \frac{1}{4} {x_1}^4\Big) {x_2}^2 \exp(-x_2).
\end{align*}

\noindent To facilitate the GP fit, the Hosaki function is subsequently standardised by its mean (0.817) and standard deviation (0.573). The noise function is

\begin{align}\label{eq:hos_noise}
g_{\text{Hosaki}}(x_1, x_2) = 50 \cdot \frac{1}{(x_1 - 3.5)^2 + 2.5}\cdot \frac{1}{(x_2 - 2)^2 + 2.5}.
\end{align}

\noindent The logarithmic Goldstein-Price function \cite{2013_Picheny} is

\begin{align}\label{eq_g-p}
    \text{G-P}(x_1, x_2) = \frac{1}{2.427} \Bigg[\log \Big([1 + {(\bar{x}_{1} + \bar{x}_{2} + 1)}^2 (19 - 14\bar{x}_{1} + 3{\bar{x}_{1}}^2 - 14\bar{x}_{2} + 6\bar{x}_{1} \bar{x}_{2} + 3{\bar{x}_{2}}^2)] \\ [30 + {(2\bar{x}_{1} - 3\bar{x}_{2})}^2 (18 - 32\bar{x}_{1} + 12{\bar{x}_{1}}^2 + 48\bar{x}_{2} - 36\bar{x}_{1} \bar{x}_{2} + 27{\bar{x}_{2}}^2)]\Big) - 8.693\Bigg]
\end{align}

\noindent where $\bar{x}_1 = 4x_1 - 2 $ and $\bar{x}_2 = 4x_2 - 2 $. The Goldstein-Price noise function is

\begin{align}\label{eq_g-p_noise}
   g_{\text{G-P}}(x_1, x_2) = \frac{3}{2} \cdot \frac{1}{(x_1 - 0.5)^2 + 0.2} \cdot \frac{1}{(x_2 - 0.3)^2 + 0.3}.
\end{align}

\begin{figure*}
\centering
\subfigure[Latent Function]{\label{fig:hos_1}\includegraphics[width=0.32\textwidth]{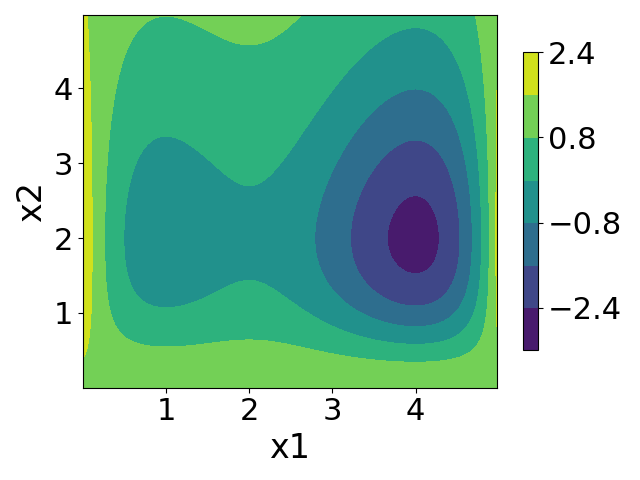}}
\subfigure[Noise Function ]{\label{fig:hos_2}\includegraphics[width=0.32\textwidth]{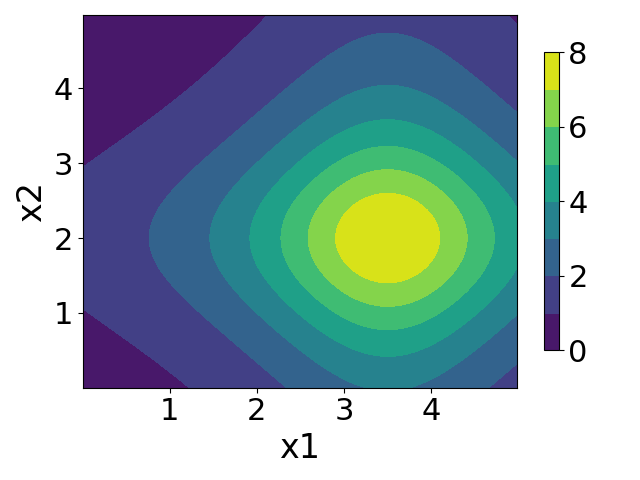}}
\subfigure[Objective Function]{\label{fig:hos_3}\includegraphics[width=0.32\textwidth]{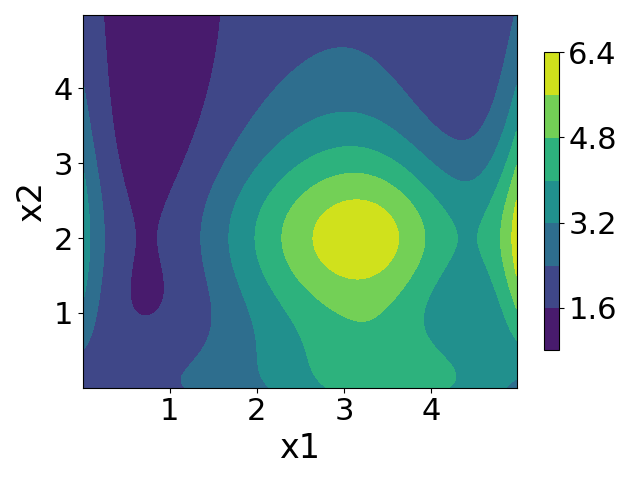}} 
\caption{(a) The latent Hosaki Function $f(\mathbf{x})$ together with (b) its heteroscedastic noise function $g(\mathbf{x})$ and (c) the objective function $f(\mathbf{x}) + g(\mathbf{x})$.}
\label{fig:hos_diagram}
\end{figure*}

\noindent For clarity, only the results of the Hosaki function are presented in the main paper with the Branin-Hoo and Goldstein-Price results presented in \ref{more_ablation}. The Hosaki function is visualised in \autoref{fig:hos_diagram}. The value of $\beta$ for ANPEI is set to 0.5 and the value of $\gamma$ is set to 500 for all Hosaki function experiments.

\subsubsection{Noiseless Case}

\begin{figure*}[t]
\centering
    \includegraphics[width=.7\textwidth]{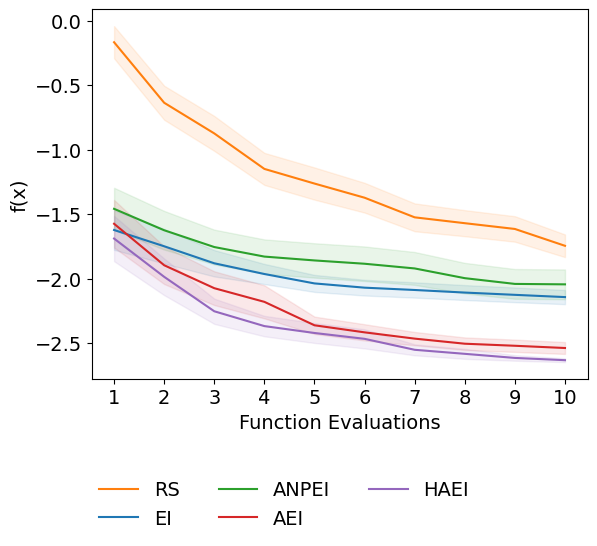}
    \caption{Hosaki function noiseless case. All Bayesian optimisation methods outperform random search. HAEI performs best and ANPEI performs worst.}
    \label{noiseless_hos}
\end{figure*}

In this case, the synthetic functions do not possess any observation noise and the optimisation function corresponds to the situation in \autoref{fig:hos_1}. 9 points sampled uniformly at random are used for initialisation and the results are displayed in \autoref{noiseless_hos}. As expected, all Bayesian optimisation methods outperform random search in the noiseless case. In this example it is unclear as to whether heteroscedastic Bayesian optimisation methods are detrimental as HAEI performs best whereas ANPEI performs worst.

\subsubsection{Homoscedastic Noise Case}

\begin{figure*}[t]
\centering
    \includegraphics[width=.7\textwidth]{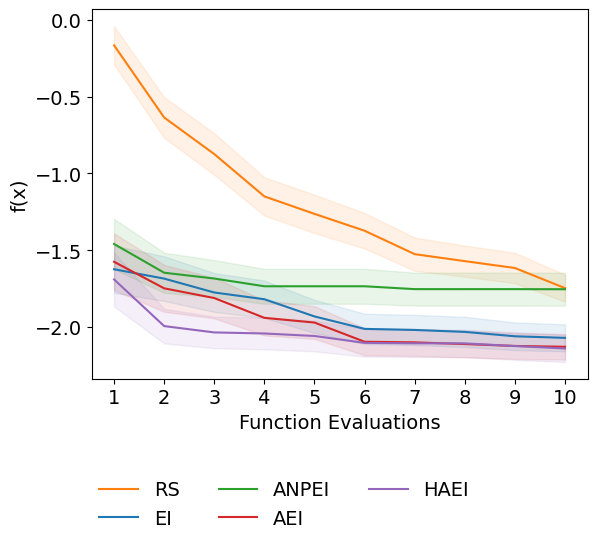}
    \caption{Hosaki function homoscedastic noise case. All Bayesian optimisation methods outperform random search with HAEI the best and ANPEI the worst.}
    \label{homo_hos}
\end{figure*}

In this case the functions are subject to homoscedastic noise of the form $25\epsilon$ where epsilon is noise sampled from a standard Gaussian $\mathcal{N}(0, 1)$. The GP surrogates are again initialised with 9 points. The results are displayed in \autoref{homo_hos}. The Bayesian optimisation methods perform worse in the homoscedastic noise case relative to the noiseless case although the rank order of the methods mirrors that of the noiseless case.

\subsubsection{Heteroscedastic Noise}

\begin{figure*}
\centering
\subfigure[Best Objective Value Found so Far]{\label{fig:bo_hosa}\includegraphics[width=0.498\textwidth]{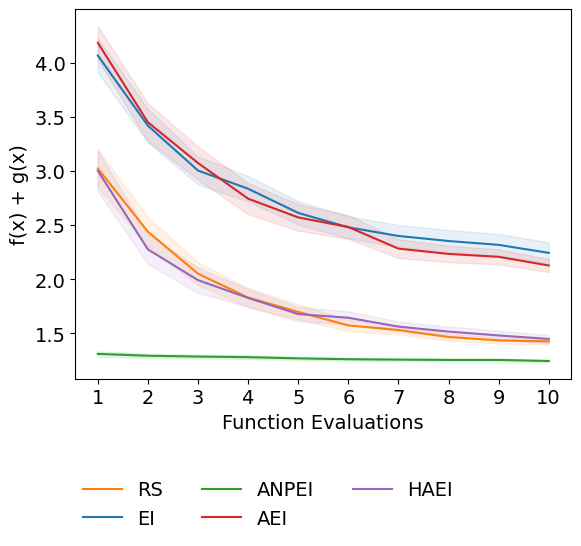}}
\subfigure[Lowest Aleatoric Noise Found so Far ]{\label{fig:bo_2_hosa}\includegraphics[width=0.482\textwidth]{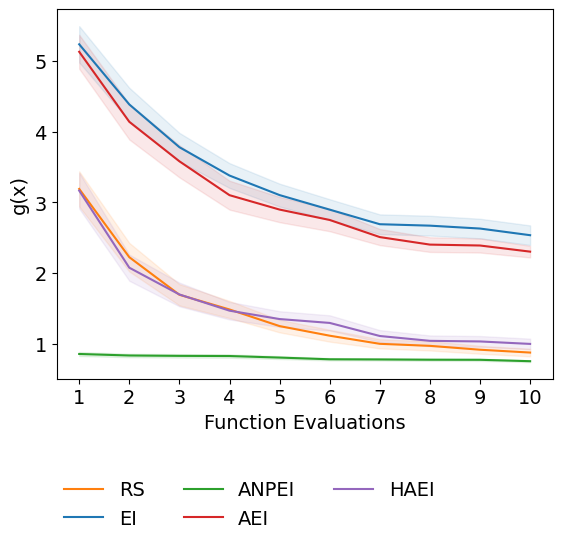}}
\caption{Comparison of heteroscedastic and homoscedastic Bayesian optimisation on the heteroscedastic 2D Hosaki function. (a) shows the optimisation of $h(\boldsymbol{x}) = f(\boldsymbol{x}) + g(\boldsymbol{x})$ (lower is better) where $g(\boldsymbol{x})$ is the aleatoric noise. (b) shows the values $g(\boldsymbol{x})$ obtained over the course of the optimisation of $h(\boldsymbol{x})$.}
\label{fig:hosaki_hetero}
\end{figure*}

In the heteroscedastic noise case the Hosaki function is subject to the noise function given in \autoref{eq:hos_noise} and visualised in \autoref{fig:hos_diagram}. 144 points were used to initialise the GP surrogates. The results are shown in \autoref{fig:hosaki_hetero}. In this instance, given that the extent of heteroscedastic noise is very strong (relative to the homoscedastic noise case), random search is highly competitive with the Bayesian optimisation methods. ANPEI however, is the best-performing algorithm. The large number of initialisation points chosen for this experiment reflects one limitation of the heteroscedastic surrogate approach; for the MLHGP to effectively learn a decomposition of the function into signal and noise components it needs access to more samples. As such, this merits an investigation into the effect of the number of samples on the performance of the heteroscedastic acquisitions.

\subsection{Effect of Initialisation Set Size}

\begin{figure*}
\centering
\subfigure[9 Points]{\label{fig:init_grid_1}\includegraphics[width=0.315\textwidth]{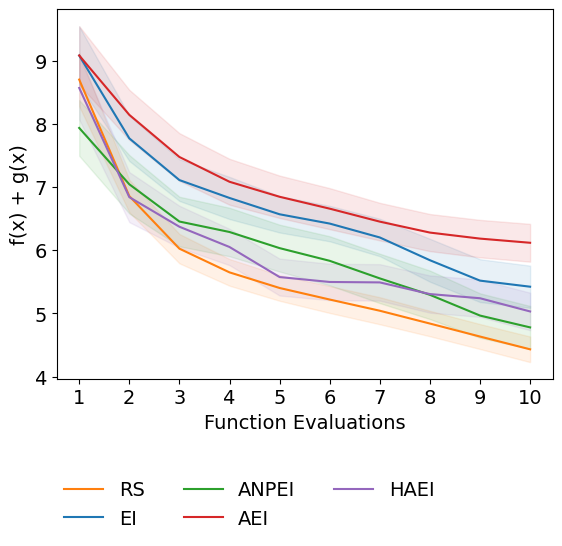}}
\subfigure[49 Points]{\label{fig:init_grid_2}\includegraphics[width=0.32\textwidth]{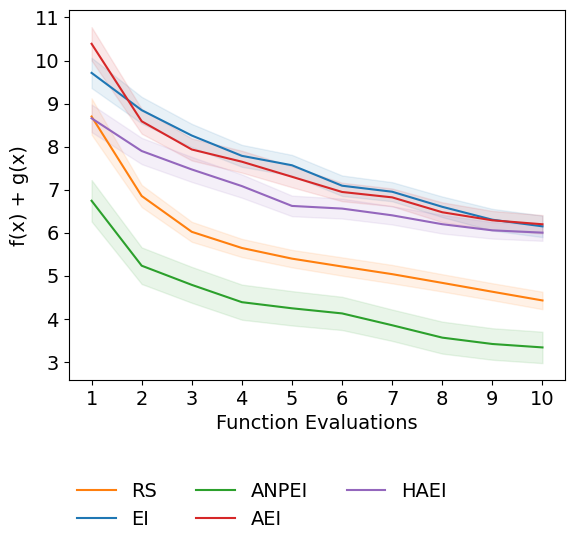}}
\subfigure[100 Points]{\label{fig:init_grid_3}\includegraphics[width=0.32\textwidth]{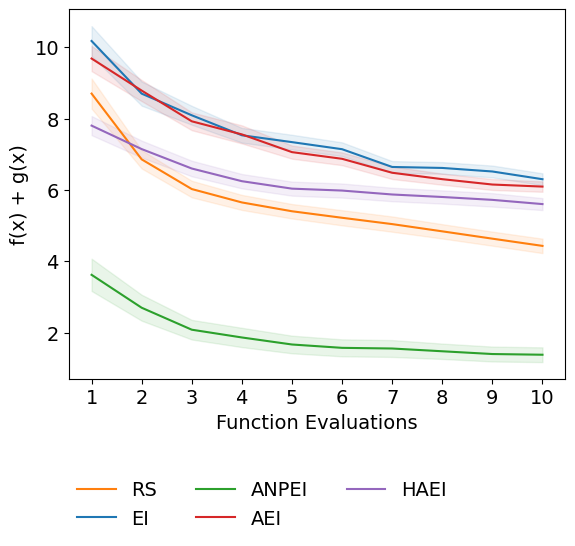}} 
\caption{The effect of the initialisation set size on the heteroscedastic Branin-Hoo function. The performance of heteroscedastic acquisitions ANPEI and HAEI increases as they are given access to more samples. An excess of samples do not help the homoscedastic Bayesian optimisation methods as they are unable to model the heteroscedastic noise component.}
\label{fig:init_grid}
\end{figure*}

The effect of the size of the initialisation set on the heteroscedastic Branin-Hoo task is investigated in \autoref{fig:init_grid}. The value of $\beta$ used for ANPEI is $\frac{1}{11}$ and the value of $\gamma$ used for HAEI is 500. The performance of the heteroscedastic acquisitions ANPEI and HAEI is observed to improve as the size of the initialisation set increases. In contrast, the homoscedastic methods EI and AEI do not improve on obtaining access to more samples as they are unable to model the heteroscedastic noise component of the task.

\subsection{Conclusions from Ablation Experiments}

Synthesising the results from the additional ablation experiments in \ref{more_ablation} some trends may be observed:

\begin{enumerate}
    \item All Bayesian optimisation methods outperform random search in the noiseless case and homoscedastic noise cases on aggregate across the three synthetic functions.
    \item On aggregate there is no significant difference between Bayesian optimisation methods in the noiseless or homoscedastic noise cases (HAEI marginally outperforms ANPEI on 2/3 noiseless tasks and 2/3 homoscedastic noise tasks).
    \item The heteroscedastic acquisitions ANPEI and HAEI perform competitively on the noiseless and homoscedastic noise tasks most likely because the MLHGP is capable of effecting nonstationary behaviour by "fantasising" heteroscedastic noise. As such, the MLHGP surrogate may be achieving enhanced flexibility relative to the homoscedastic GP in this setting.
    \item The heteroscedastic acquisitions tend to outperform other Bayesian optimisation approaches on the heteroscedastic noise tasks although crucially this depends on the size of the initialisation set. In order to detect heteroscedastic noise the MLHGP surrogate needs access to more samples relative to the noiseless and homoscedastic cases.
    \item ANPEI outperforms HAEI.
\end{enumerate}

In summary, the experiments would appear to show that there is no significant downside to employing a heteroscedastic surrogate and acquisition function on noiseless tasks or tasks with homoscedastic noise save for the increased training time for the model.

\section{Conclusions}

We have presented an approach for performing Bayesian optimisation with the explicit goal of minimising aleatoric noise in the suggestions. We posit that such an approach can prove useful for the natural sciences in the search for molecules and materials that are robust to experimental measurement noise. The synthetic function ablation study highlights no particular downside to the use of the MLHGP in conjunction with ANPEI or HAEI in cases where the noise structure of the problem is a priori unknown i.e the black-box optimisation problem is either noiseless or homoscedastic. Nonetheless, we anticipate that this type of approach may be particularly relevant for the experimental natural sciences where noiseless objectives or those with homoscedastic noise are highly uncommon. In terms of concrete recommendations on when to apply the algorithm, we foresee the best performance in situations where the user has access to a moderately-sized initialisation set in order to provide the MLHGP with enough samples to distinguish heteroscedastic noise from intrinsic function variability. There are a number of possible extensions to the current approach which may facilitate its application to high-dimensional datasets and act as fruitful sources for future work:

\begin{enumerate}[label=(\arabic*)]
    \item \textbf{Surrogate Model:} One disadvantage of the MLHGP model is the lack of convergence guarantees for the EM-like procedure required for fitting. Various other forms of heteroscedastic GP exist \cite{2005_Le, 2018_Binois, 2017_Almosallam, 2011_Munoz, 2017_Wang, 2012_Wang, 2020_Zhang} and have demonstrated success in modelling applications \cite{2018_Rodrigues, 2018_Tabor, 2020_Rogers, 2019_Wang}. Of particular interest for real-world problems are scalable heteroscedastic GPs \cite{2019_Wang_gp, 2020_Liu} which could circumvent the computationally-intensive bottleneck of fitting multiple exact GPs as a subroutine of the MLHGP Bayesian optimisation procedure.
    \item \textbf{Advances in Surrogate Model Machinery}: Advances in areas such as efficient sampling of GPs \cite{2020_Wilson} are liable to yield improvements to sampled-based acquisition functions such as Thompson sampling \cite{1933_Thompson} while fully Bayesian approaches to hyperparameter estimation for sparse GPs \cite{2019_Lalchand} are liable to yield improvements in model fitting procedures.
    \item \textbf{Scalable Bayesian Optimisation:} Scalable Bayesian optimisation can also be enabled via dimensionality reduction techniques \cite{2020_Moriconi, 2019_Candelieri, 2021_Grosnit}. Such approaches, when combined with efficient libraries \cite{2019_Balandat, 2019_dragonfly} could facilitate heteroscedastic Bayesian optimisation in high-dimensional settings.
    \item \textbf{Acquisition Function Optimisation:} Recent developments in acquisition function optimisation including Monte Carlo reformulations \cite{2018_Wilson, 2020_Grosnit}, compositional optimisers \cite{2020_Tutunov, 2020_Grosnit} and tight relaxations \cite{2020_Schweidtmann} of common acquisition functions have the potential to yield gains in empirical performance.
    \item \textbf{Data Transformation:} Input-warping \cite{2020_Wiebe} and output transformations \cite{2020_Rivers} have recently shown success towards addressing heteroscedastic datasets \cite{2021_Mrk}.
    \item \textbf{Approaches for Molecular Bayesian Optimisation:} In relation to molecules, the use of tailored GP kernels such as Tanimoto kernels \cite{2020_Moss, 2020_Thawani} and more expressive dimensionality reduction techniques \cite{2020_Cheng} could lead to performance gains and enhanced scalability respectively.
    \item \textbf{Exploration in the Noise Objective:} Incorporating exploration in the noise objective in the multi-objective setting as in \cite{2013_Kuindersma}.
\end{enumerate}

\noindent Lastly, a further use-case of the machinery developed in this paper is obtained by turning the noise minimisation problem into a noise maximisation problem. As an example, in materials discovery, we may derive benefit from being antifragile \cite{2012_taleb} towards (i.e. derive benefit from) high aleatoric noise. In an application such as the search for performant perovskite solar cells, we are faced with an extremely large compositional space, with millions of potential candidates possessing high aleatoric noise for identical reproductions \cite{2019_Zhou}. In this instance we may want to guide search towards a candidate possessing a high photoluminescence quantum efficiency with high aleatoric noise. If the cost of repeating material syntheses is small relative to the cost of the search, the large aleatoric noise will present opportunities to synthesise materials possessing efficiencies far in excess of their mean values.

\section{Acknowledgements}

The authors would like to thank James T. Wilson for discussion about the experimental setup and applications domains of heteroscedastic Bayesian optimisation, Henry Moss for useful discussions regarding Bayesian optimisation in the noisy setting and Luke Corcoran for discussions relating to the limiting behaviour of the HAEI acquisition. We would additionally like to thank the anonymous reviewers who were instrumental in improving the quality of the empirical analysis as well as the clarity of the manuscript. VL is funded by The Alan Turing Institute Doctoral Studentship under the EPSRC grant EP/N510129/1.

\newpage

\bibliography{bib.bib}

\newpage

\appendix

\section{Heteroscedasticity of the Soil Phosphorus Fraction Dataset}

\autoref{the_table} is used to demonstrate the efficacy of modelling the soil phosphorus fraction dataset using a heteroscedastic GP. The heteroscedastic GP outperforms the homoscedastic GP on prediction based on the metric of negative log predictive density (NLPD)

\begin{equation*}
    \text{NLPD} = \frac{1}{n} \sum_{i=1}^n - \log p(t_i | \boldsymbol{x_i})
\end{equation*}

\noindent which penalises both over and under-confident predictions.

\begin{table}[ht]
\centering
\caption{Comparison of NLPD values on the soil phosphorus fraction dataset. Standard errors are reported for 10 independent train/test splits. Lower scores are better.}
\label{the_table}
\vspace{3mm}
\resizebox{0.75\textwidth}{!}{%
\begin{tabular}{@{}lll@{}}
\toprule
\textbf{Soil Phosphorus Fraction Dataset} & \multicolumn{1}{c}{\textbf{GP}} & \multicolumn{1}{c}{\textbf{Het GP}} \\ \midrule
NLPD & $1.35 \pm 1.33$ & $1.00 \pm 0.95$ \\ \bottomrule
\end{tabular}%
}
\end{table}

\section{Additional Ablation Experiments}\label{more_ablation}

In this section we present the ablation results on noiseless, homoscedastic and heteroscedastic noise tasks in line with \autoref{first_ablation} of the main paper.

\subsection{Goldstein-Price Function}

\begin{figure*}
\centering
\subfigure[Latent Function]{\label{fig:gp_1}\includegraphics[width=0.32\textwidth]{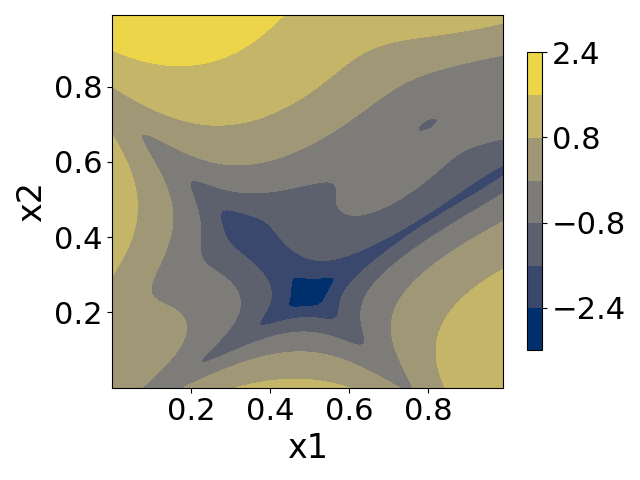}}
\subfigure[Noise Function ]{\label{fig:gp_2}\includegraphics[width=0.32\textwidth]{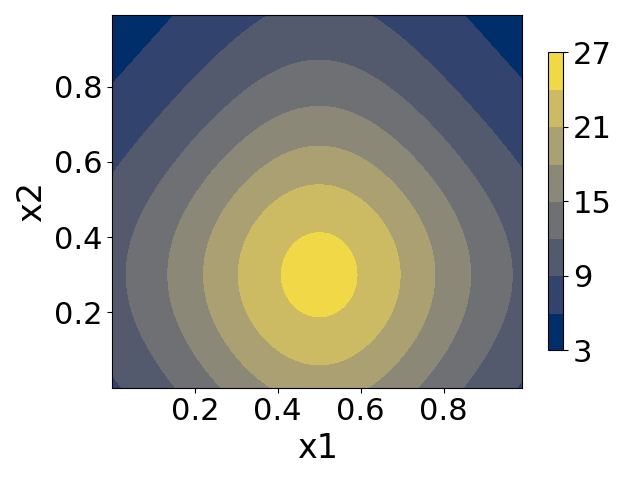}}
\subfigure[Objective Function]{\label{fig:gp_3}\includegraphics[width=0.32\textwidth]{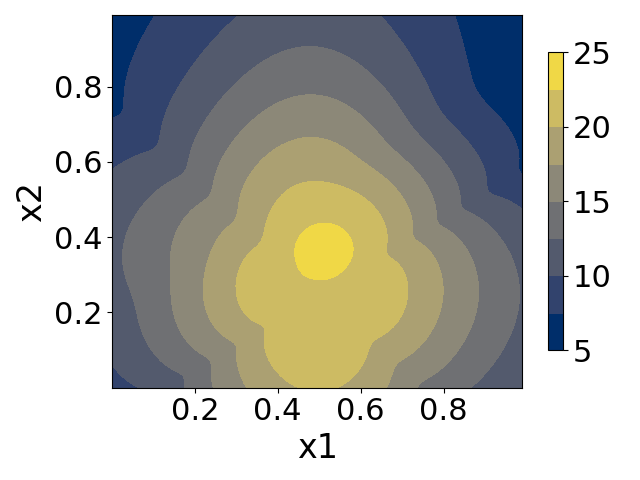}} 
\caption{(a) The latent Goldstein-Price Function $f(\mathbf{x})$ together with (b) its heteroscedastic noise function $g(\mathbf{x})$ and (c) the objective function $f(\mathbf{x}) + g(\mathbf{x})$..}
\label{fig:g-p_function}
\end{figure*}

The form of the Goldstein-Price function is given in \autoref{eq_g-p} with noise function in \autoref{eq_g-p_noise}. The function is visualised in \autoref{fig:g-p_function}. 9 data points are used for initialisation in the noiseless and homoscedastic noise cases whereas 100 data points are used for initialisation in the heteroscedastic noise case. $\beta$ is set to 0.5 for the noiseless and homoscedastic noise tasks and $\frac{1}{11}$ for the heteroscedastic noise task. $\gamma$ is set to 500 for all experiments.

\subsubsection{Noiseless Case}

\begin{figure*}[t]
\centering
    \includegraphics[width=.7\textwidth]{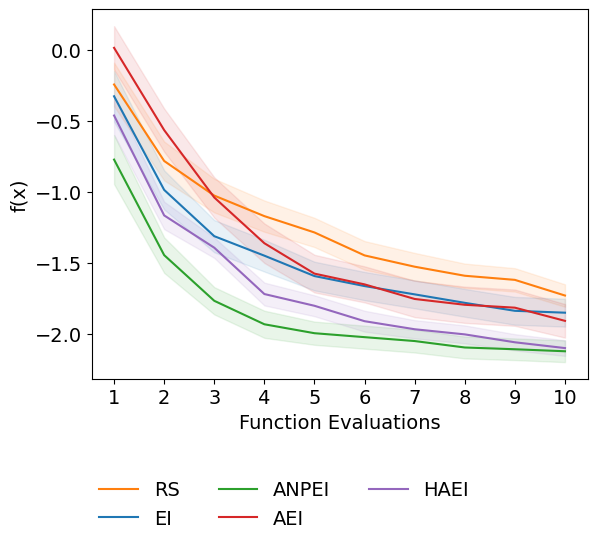}
    \caption{Goldstein-Price function noiseless case. All Bayesian optimisation methods outperform random search. ANPEI performs best and HAEI is runner-up.}
    \label{noiseless_g-p}
\end{figure*}

The results of the noiseless case for Goldstein-Price are given in \autoref{noiseless_g-p}. All Bayesian optimisation methods outperform random search with ANPEI best and HAEI second best.

\pagestyle{fancy}
\fancyhf{}
\lhead{Appendix}
\rhead{\thepage}

\subsubsection{Homoscedastic Noise Case}

\begin{figure*}[t]
\centering
    \includegraphics[width=.7\textwidth]{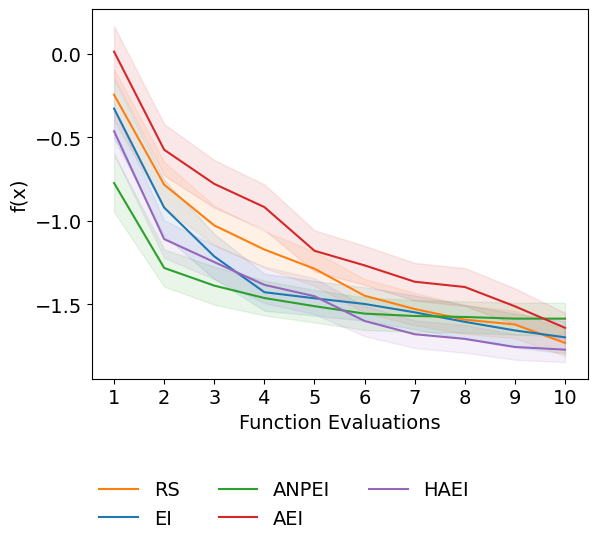}
    \caption{Goldstein-Price function homoscedastic noise case. HAEI performs best.}
    \label{homo_g-p}
\end{figure*}

The results of the homoscedastic noise case for Goldstein-Price are shown in \autoref{homo_g-p}. In this instance HAEI performs best.

\subsubsection{Heteroscedastic Noise}

\begin{figure*}
\centering
\subfigure[Best Objective Value Found so Far]{\label{fig:bo_g-p}\includegraphics[width=0.49\textwidth]{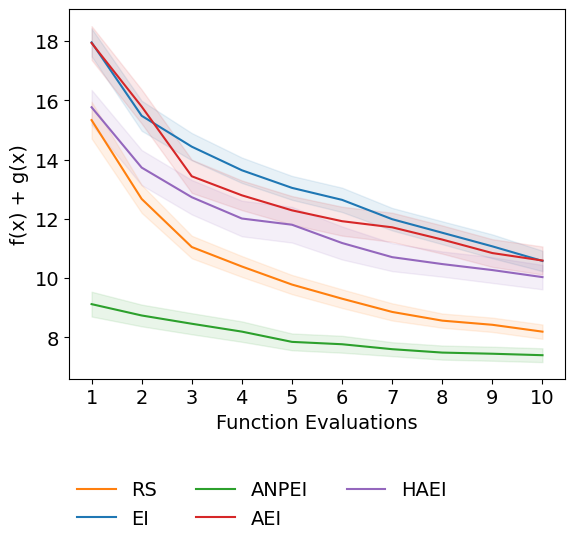}}
\subfigure[Lowest Aleatoric Noise Found so Far ]{\label{fig:bo_2_g-p}\includegraphics[width=0.49\textwidth]{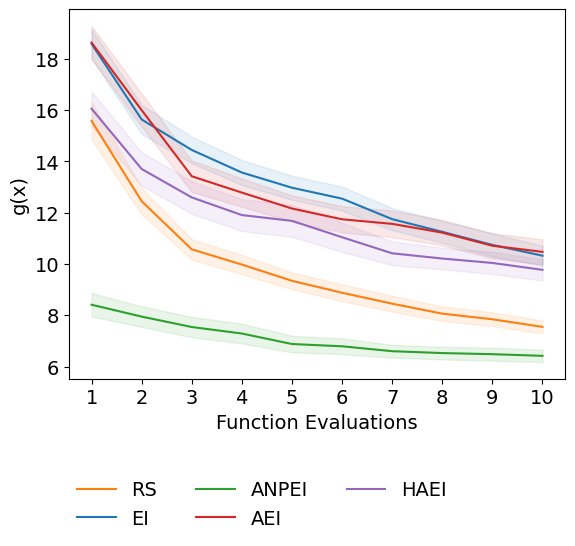}}
\caption{Comparison of heteroscedastic and homoscedastic Bayesian optimisation on the heteroscedastic 2D Goldstein-Price function. (a) shows the optimisation of $h(\boldsymbol{x}) = f(\boldsymbol{x}) + g(\boldsymbol{x})$ (lower is better) where $g(\boldsymbol{x})$ is the aleatoric noise. (b) shows the values $g(\boldsymbol{x})$ obtained over the course of the optimisation of $h(\boldsymbol{x})$.}
\label{fig:g-p_hetero}
\end{figure*}

The results of the heteroscedastic noise case for Goldstein-Price are shown in \autoref{fig:g-p_hetero}. ANPEI performs best whilst HAEI performs worse than random search.

\subsection{Branin-Hoo Function}

\begin{figure*}
\centering
\subfigure[Latent Function]{\label{fig:double_bran_1}\includegraphics[width=0.32\textwidth]{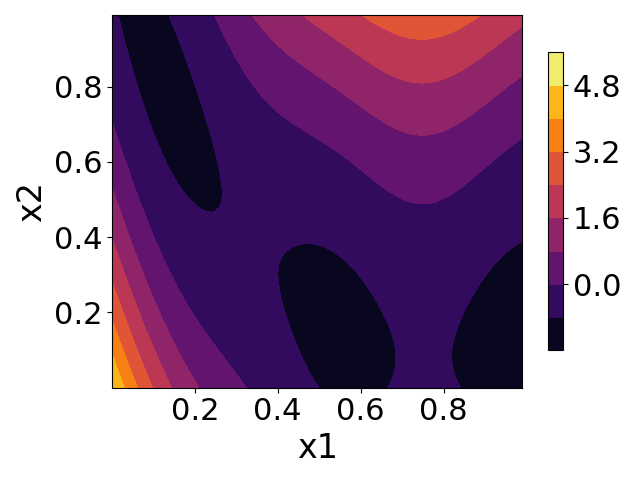}}
\subfigure[Noise Function ]{\label{fig:double_bran_2}\includegraphics[width=0.32\textwidth]{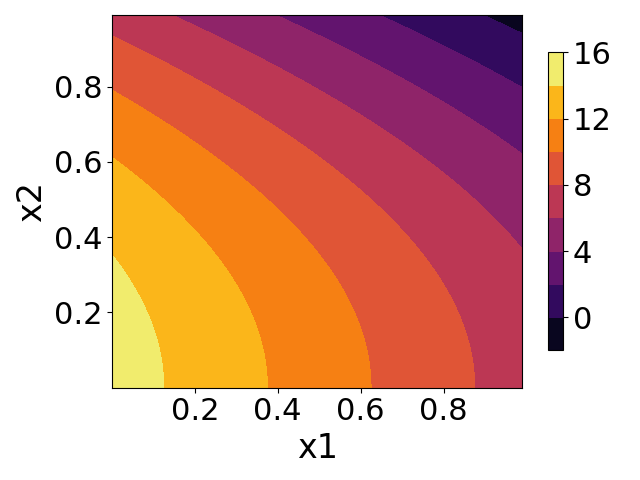}}
\subfigure[Objective Function]{\label{fig:double_bran_3}\includegraphics[width=0.32\textwidth]{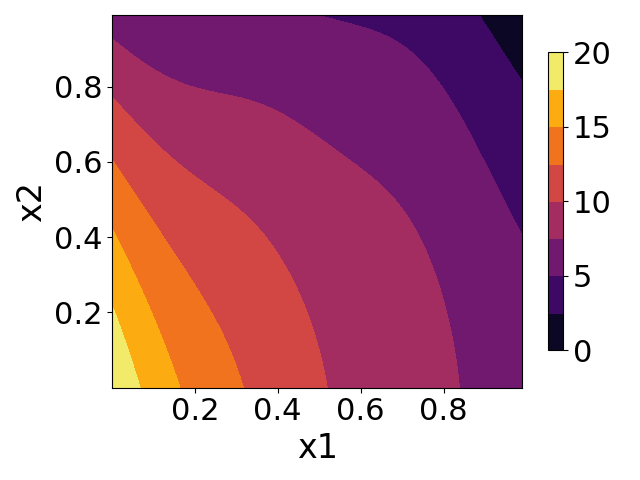}} 
\caption{Heteroscedastic Branin Function.}
\label{fig:double_bran}
\end{figure*}

The form of the Branin-Hoo function is given in \autoref{branin_eq} with noise function in \autoref{branin_noise_eq}. The function is visualised in \autoref{fig:double_bran}, a figure from the main paper repeated here for clarity. 9 data points are used for initialisation in the noiseless and homoscedastic noise cases whereas 100 data points are used for initialisation in the heteroscedastic noise case. $\beta$ is set to 0.5 and $\gamma$ is set to 500 for all experiments.

\subsubsection{Noiseless Case}

\begin{figure*}[t]
\centering
    \includegraphics[width=.7\textwidth]{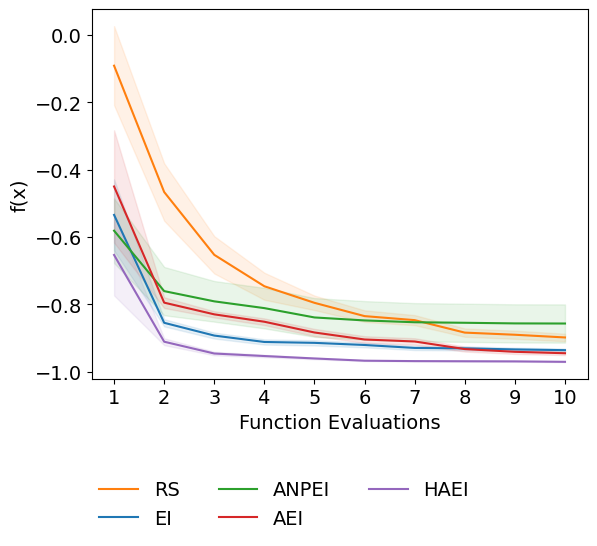}
    \caption{Branin-Hoo function noiseless case. HAEI performs best. ANPEI performs worst.}
    \label{noiseless_double_branin}
\end{figure*}

The results of the noiseless case for the Branin-Hoo function are given in \autoref{noiseless_double_branin}. HAEI performs best in this case whereas ANPEI performs worst.

\subsubsection{Homoscedastic Noise Case}

\begin{figure*}[t]
\centering
    \includegraphics[width=.7\textwidth]{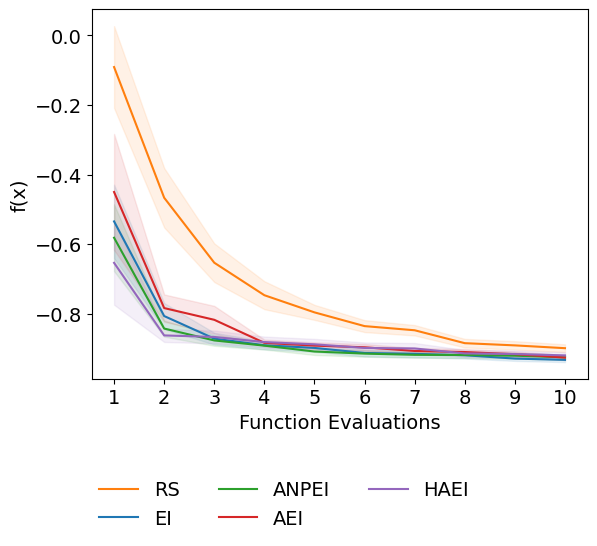}
    \caption{Branin-Hoo function homoscedastic noise case. All Bayesian optimisation methods outperform random search.}
    \label{homoscedastic_braninhoo}
\end{figure*}

The results of the homoscedastic noise case for the Branin-Hoo function are given in \autoref{homoscedastic_braninhoo}. All Bayesian optimisation methods outperform random search yet perform comparably against each other.

\subsubsection{Heteroscedastic Noise}

\begin{figure*}
\centering
\subfigure[Best Objective Value Found so Far]{\label{fig:bo_bran2}\includegraphics[width=0.495\textwidth]{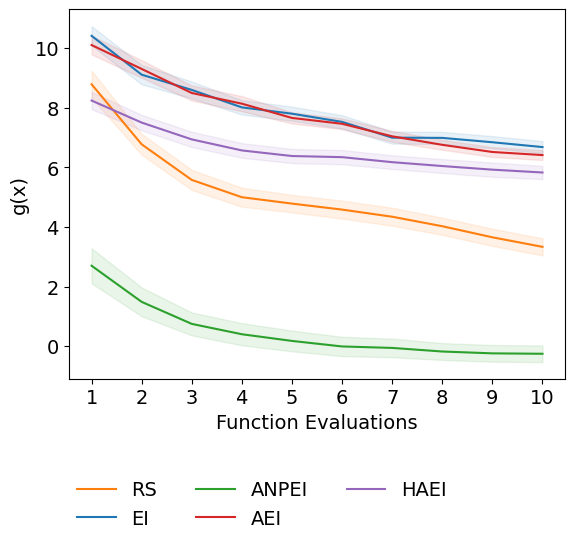}}
\subfigure[Lowest Aleatoric Noise Found so Far ]{\label{fig:bo_2_bran2}\includegraphics[width=0.493\textwidth]{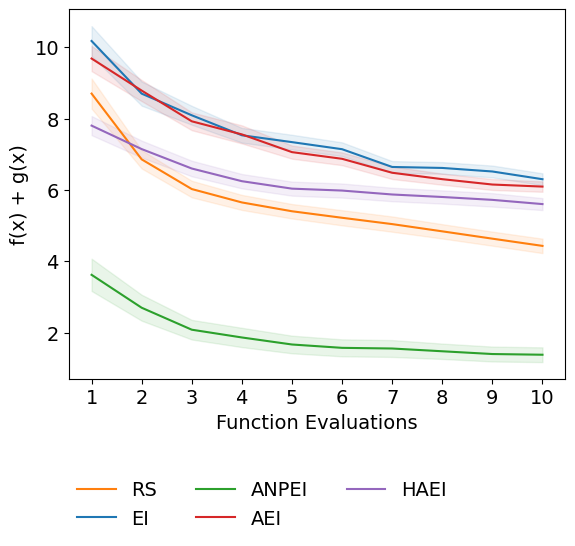}}
\caption{Comparison of heteroscedastic and homoscedastic Bayesian optimisation on the heteroscedastic 2D Branin function. (a) shows the optimisation of $h(\boldsymbol{x}) = f(\boldsymbol{x}) + g(\boldsymbol{x})$ (lower is better) where $g(\boldsymbol{x})$ is the aleatoric noise. (b) shows the values $g(\boldsymbol{x})$ obtained over the course of the optimisation of $h(\boldsymbol{x})$.}
\label{fig:bran_hetero2}
\end{figure*}

The results of the heteroscedastic noise case for the Branin-Hoo function are shown in \autoref{fig:bran_hetero2}. ANPEI performs best whilst HAEI performs worse than random search.

\section{Performance Impact of the Kernel Choice}\label{kernel_exps}

\begin{figure*}
\centering
\subfigure[ANPEI]{\label{fig:bo_brank}\includegraphics[width=0.495\textwidth]{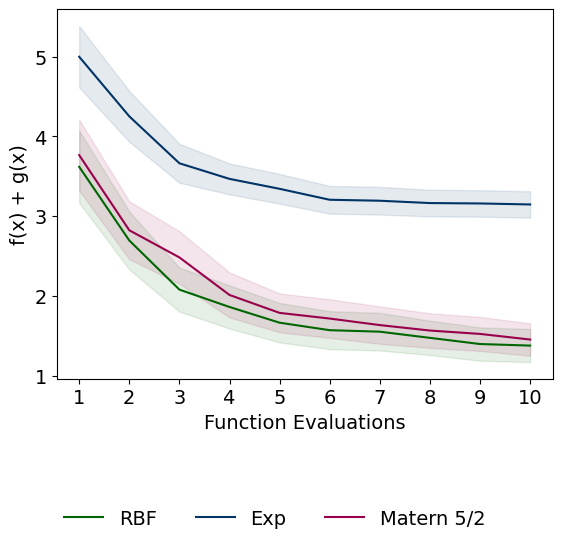}}
\subfigure[HAEI]{\label{fig:bo_2_brank}\includegraphics[width=0.495\textwidth]{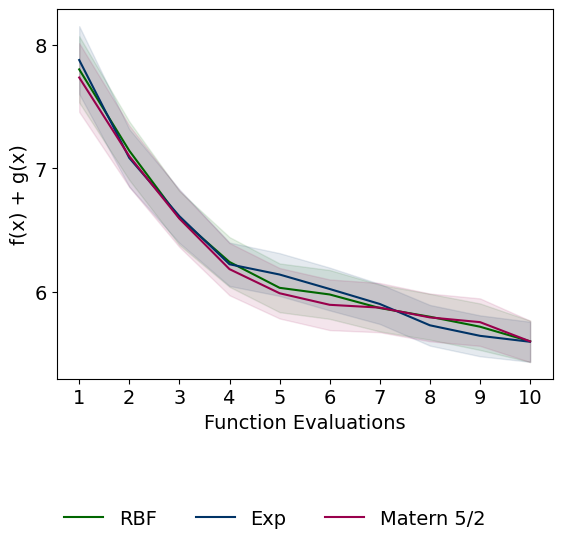}}
\caption{Branin-Hoo function kernel comparison.}
\label{fig:bran_kernel}
\end{figure*}

\begin{figure*}
\centering
\subfigure[ANPEI]{\label{fig:bo_g-p}\includegraphics[width=0.486\textwidth]{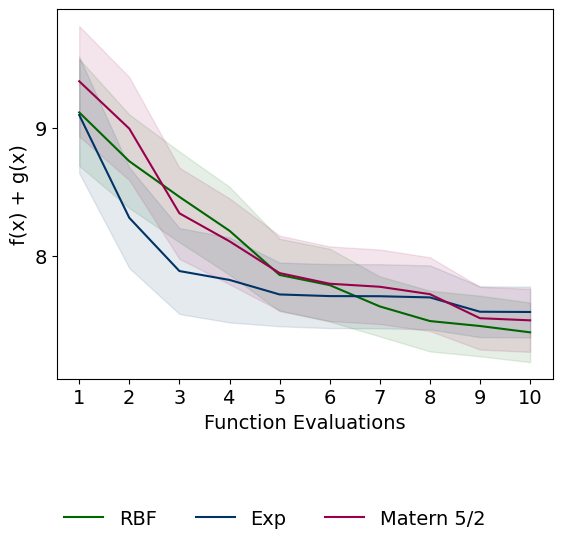}}
\subfigure[HAEI]{\label{fig:bo_2_g-p}\includegraphics[width=0.495\textwidth]{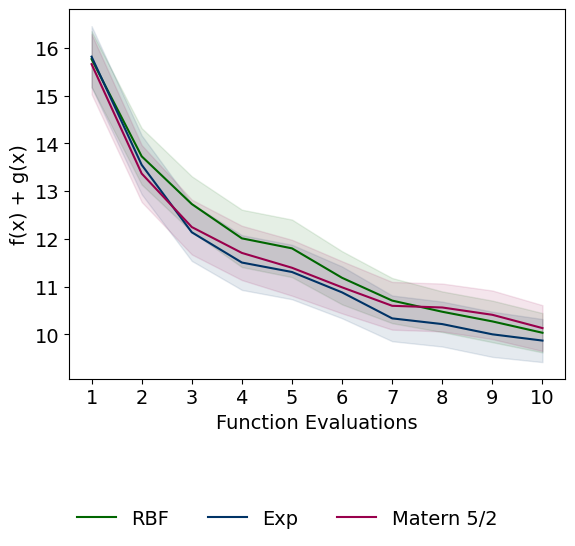}}
\caption{Goldstein-Price function kernel comparison.}
\label{fig:gold_kernel}
\end{figure*}

\begin{figure*}
\centering
\subfigure[ANPEI]{\label{fig:bo_hosk}\includegraphics[width=0.495\textwidth]{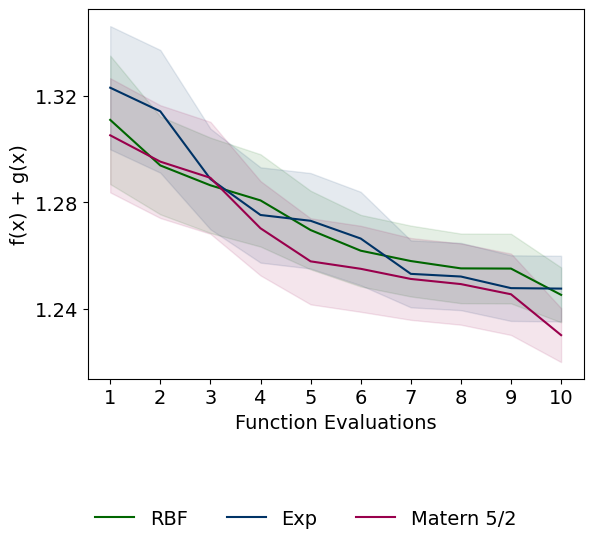}}
\subfigure[HAEI]{\label{fig:bo_2_hosk}\includegraphics[width=0.495\textwidth]{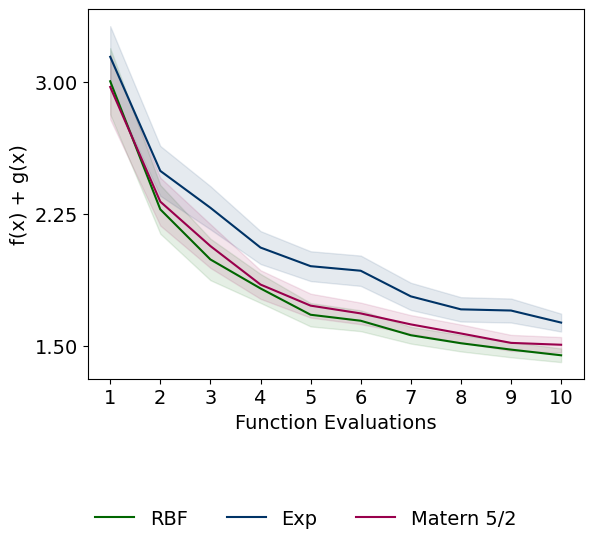}}
\caption{Hosaki Function kernel comparison.}
\label{fig:hos_kernel}
\end{figure*}

In this section we analyse the impact that the choice of GP kernel has on Bayesian optimisation performance. We select three kernels for this purpose: the squared exponential kernel

\begin{equation*}
    k_{\text{SQE}}(\boldsymbol{x}, \boldsymbol{x'}) = \sigma_{f}^{2}\cdot\text{exp}\Big(\frac{-\lVert\boldsymbol{x} - \boldsymbol{x'}\rVert^{2}}{2\ell^{2}}\Big)
\end{equation*}

\noindent used for all experiments in the main paper, the exponential kernel

\begin{equation*}
    k_{\text{exp}}(\boldsymbol{x}, \boldsymbol{x'}) = \sigma_{f}^{2}\cdot\text{exp}\Big(\frac{-\lVert\boldsymbol{x} - \boldsymbol{x'}\rVert}{\ell}\Big),
\end{equation*}

\noindent a special instance of the Mat\'{e}rn kernel for values of $\nu = \frac{1}{2}$ \cite{2006_Rasmussen} as well as the Mat\'{e}rn 5/2 kernel

\begin{equation*}
    k_{\text{Mat\'{e}rn}(5/2)}(\boldsymbol{x}, \boldsymbol{x'}) = \sigma_{f}^{2}\cdot \Big(1 + \frac{\sqrt{5} \lVert\boldsymbol{x} - \boldsymbol{x'}\rVert}{\ell} + \frac{5 \lVert\boldsymbol{x} - \boldsymbol{x'}\rVert^2}{3\ell^2}\Big) \cdot \text{exp}\Big(\frac{-\sqrt{5} \lVert\boldsymbol{x} - \boldsymbol{x'}\rVert}{\ell}\Big)
\end{equation*}

\noindent which is one of the most popular kernels for large scale empirical studies \cite{2018_Wilson, 2020_Grosnit}. It should be noted that while the equations are written assuming a single scalar lengthscale, in practice for the experiments in greater than 1D, each lengthscale is optimised per dimension under the marginal likelihood. For all experiments we choose the same kernel for both GPs of the MLHGP model i.e. the GP modelling the objective as well as the GP modelling the noise. 100 points are used for initialisation in the Branin-Hoo and Goldstein-Price functions and 144 points are used for the Hosaki function. $\beta$ is set to 0.5 for the Branin-Hoo and Hosaki functions and $\frac{1}{11}$ for the Goldstein-Price function. $\gamma$ is set to 500 for all experiments. The results are shown in \autoref{fig:bran_kernel}, \autoref{fig:gold_kernel} and \autoref{fig:hos_kernel} for the Branin-Hoo function, Goldstein-Price function and Hosaki functions respectively. There is no significant difference in performance using each kernel save for the Branin-Hoo function where ANPEI underperforms using the somewhat rougher exponential kernel.

\end{document}